\documentclass[11pt]{article}
\usepackage[utf8]{inputenc}
\usepackage[T1]{fontenc}
\usepackage{amsmath,amssymb,amsfonts}
\usepackage[margin=1.15in]{geometry}
\usepackage{hyperref}
\hypersetup{colorlinks=true,linkcolor=black,citecolor=black,urlcolor=blue}
\title{Innovation-Residual Auditing of Autonomous Analysis Agents:\\
Localization, Detection Limits, Error Control, and Identifiability}
\author{
  Ahmed Hassoon\\ Johns Hopkins University\\ \texttt{ahassoo1@jhu.edu}
  \and
  Mark Dredze\\ Johns Hopkins University\\ \texttt{mdredze1@jh.edu}
}
\date{\today}
\begin{document}
\maketitle

\begin{abstract}

Autonomous agents now carry out entire data analyses, selecting cohorts, joining tables, and fitting models with little step-by-step supervision. When such an analysis turns out to be wrong, someone must determine which operation caused it. A recent approach does this without any labelled mistakes, learning instead from analyses known to be sound and flagging operations that depart from what that model predicts; how reliable such audits are has not been studied. This paper supplies that analysis. The choice of score determines whether an error can be localized at all. If each operation is scored by how surprising it is given the operation immediately preceding it, then operations that merely inherit an earlier error are indistinguishable from correct ones, so one mistake produces one flag; scores computed against a longer reconstruction of the intended analysis instead spread a single mistake across many operations. We quantify how far they spread, and how to choose the comparison length when an error accumulates gradually rather than at once. We then give procedures that control the proportion of falsely flagged operations within a single audited analysis, requiring only that sound analyses be exchangeable rather than that the fitted model be correct, and we quantify how much the guarantees weaken when the model is imperfect or when the analysis was selected for review in a way that depends on its content. Finally we establish a limit on what any such audit can report: errors below a certain magnitude cannot be attributed at all, being indistinguishable from ordinary variation among sound analyses. This limit falls so slowly as more sound analyses are collected that at the representation sizes now in use a hundredfold increase reduces it by under two percent, so the dimension of the representation rather than the volume of training data is the binding constraint.

\end{abstract}
\section{Introduction}

Language-model agents are increasingly used to conduct data analysis end to end. Given a research question, such an agent formulates hypotheses, defines cohorts, joins tables across federated repositories, fits models, and reports conclusions, executing dozens of operations without human review of the intermediate steps. In biomedical settings the resulting analyses inform substantive claims, and the volume at which they can be produced exceeds what expert reviewers can check line by line.

These analyses are not always sound. An agent may narrow a cohort with a filter that silently excludes a class of patients, join on a key that drops records, or fit a model omitting an adjustment the design requires. The result is a trajectory that terminates in a plausible-looking conclusion reached by invalid means. Detecting that a trajectory is unsound is one problem; the operative question for anyone who must act on the finding is a second one, namely which of its operations introduced the error. That determines what to re-run, what to correct, and whether the conclusion can be salvaged.

Attributing a failure to specific operations is difficult for a reason intrinsic to sequential execution. An error early in a trajectory corrupts the state that every subsequent operation consumes, so later operations that are themselves executed correctly nonetheless appear anomalous relative to what the analysis should have been doing. Symptoms mask the source. The difficulty compounds when a trajectory contains more than one genuine error, when the trajectory reaching the auditor has already been selected by an upstream judgment that is itself a function of the whole trajectory, and when no examples of failure are available to learn from.

Existing approaches divide on that last constraint. Supervised attribution methods learn from failure trajectories with annotated error steps, obtained by hand, by counterfactual replay, or by programmatic fault injection (Zhang et al., 2025; Feng et al., 2026). The cost of producing such annotation is substantial, and the common assumption of a single decisive error per trajectory reflects it. Yeh et al. (2026) showed that the annotation requirement can be removed altogether, by training exclusively on successful trajectories and scoring each step of a failed one by its deviation from the learned dynamics of success. The present work adopts that formulation and develops its statistical theory.

The statistical properties of attribution under this formulation have not yet been characterized. An audit that reports a set of operations raises three questions that benchmark accuracy does not answer: what fraction of the reported operations should be expected to be spurious, how large an error must be before detection is possible at all, and what a report means when the fitted model of successful execution is itself approximate. For audits that inform decisions about scientific results, these properties matter alongside accuracy.

This paper supplies that account. We analyze audits in which each operation is scored by the one-step innovation of a predictable predictive model fitted to successful trajectories. Localization is exact: conditional on its realized input state, an operation executed correctly after an earlier error has a null residual, so inherited corruption produces no signal and the masking problem above does not arise (Theorem 1). The result requires predictable conditioning and a one-step score; Sections 3.2 and 6.1 characterize which scoring rules satisfy these conditions and what remains available when they do not. The prediction horizon governs a trade-off between localization and accumulation, so errors introduced gradually, which are invisible to one-step scores, are recoverable at a matched horizon (Section 3.2). False discovery control over the flagged set is available under arbitrary within-trajectory dependence, in a form requiring only exchangeability across trajectories rather than correct specification of the fitted model, together with a sensitivity analysis reporting how much selection into the audit would be needed to remove each flag (Section 5).

Three further results concern the deployed model and the population of trajectories. Cost-normalized time is predictable only under a joint model of execution cost and operation embedding, a factorization that additionally makes cost an auditable channel (Section 6.1), and explicit conditions on off-support variance inflation are what leave operations inheriting an earlier error conservatively null rather than uncontrolled (Section 7). Pooling recurrent operation types across many trajectories lowers the detectable perturbation by a factor $\sqrt{m_c}$ (Section 10). Finally, the estimand itself is untestable without a minimum-effect threshold, and with one the smallest attributable error is bounded below by a floor that additional calibration data does not remove (Section 11).

Part I develops localization, score selection, and multiplicity control for arbitrary predictable one-step predictive families. Part II studies a Gaussian bridge implementation, including time normalization, terminal anchoring, and behavior outside the training support. Part III treats recurrence across trajectories, identifiability, and detection limits. Every result holds within stated modeling assumptions, and Section 5.6 prices deviations from them rather than treating specification as exact.

\subsection{Related work}

Zhang et al. (2025) formulated step-level failure attribution and released the Who\&When benchmark, establishing the difficulty of the task: the best reported step-level accuracy was 14.2\%. Subsequent studies used prompting pipelines or post-trained attribution models with step-level failure labels obtained by manual annotation, counterfactual replay, or programmatic fault injection. Feng et al. (2026) applied conformal prediction to produce contiguous prediction sets with finite-sample coverage. Zhang et al. (2026) formulated online auditing as detection of the earliest decisive error. These approaches assume a single decisive error per trajectory and require annotated failures.

Yeh et al.\ (2026) remove the annotation requirement. They embed each operation as a hidden-state vector from the acting model, treat the sequence as irregular observations of a continuous latent path from the query representation to a terminal goal state, fit a neural controlled differential equation to successful trajectories under squared reconstruction loss, and score each step by $\|h_t - \hat{h}(u_t)\|_2^2$. Steps are flagged by selecting the top $k$ scores or by thresholding at the $(1-\alpha)$ quantile of pooled per-step scores from held-out successful trajectories. Four elements are adopted from that work: the one-class formulation, the continuous latent path between an initial query state and a terminal goal state, the pooled conformal threshold calibrated on successful trajectories, and the representation of the result as a set of contributing steps.

Building on that formulation, the present work develops four elements of its theory.

The first is a predictable time grid. Sequential conditioning requires the time index entering the predictive law to be measurable with respect to the past. Grids normalized by realized trajectory length, a natural choice when trajectories vary in length, are functions of the complete trajectory and do not have this property; neither does budget normalization on its own, since an operation's cost is a property of that operation. Section 6.1 gives a construction that does, by modeling cost and embedding jointly, which additionally makes execution cost a second audit channel.

The second is a characterization of the scoring rule. A reconstruction score computed against a path driven by an interpolant of the observed trajectory lies between the one-step innovation and the free-running rollout. Its effective prediction horizon is a property of the fitted vector field rather than a design parameter, because the control path allows the dynamics to track the audited trajectory; Proposition 1 shows that this localization window can be measured. Two properties of such scores bear on the guarantees developed here. Because the dynamics track the observed trajectory, a corruption partly predicts itself, attenuating its own score along with those downstream. And when the interpolant is bidirectional, as smoothing and natural cubic splines are, the value at $u_t$ depends on knots after $t$, so the score at a step is a function of later steps and is not predictable in the sense Theorem 1 requires. Implementations can be checked against both conditions directly.

The third is error control over the reported set. A fixed quantile of pooled calibration scores controls a marginal false-positive rate for in-distribution steps. The fraction of false flags among the operations actually reported is a distinct quantity, and the two differ because flags within a trajectory are dependent through the shared calibration set and through the trajectory itself. Section 5 supplies the latter under that dependence, together with a sensitivity analysis for selection of trajectories into the audit population.

The fourth is a set of limits: Sections 4 and 11 derive detection boundaries, minimum detectable perturbations, and identifiability results for this formulation, which bound what any procedure within it can report.

The control path of Yeh et al. (2026) also resolves the multimodal endpoint problem of Section 8, and does so more directly than the construction developed here. A latent path determined only by its initial condition is a deterministic flow from query to endpoint and cannot represent two distinct valid conclusions, as formalized in Proposition 10. Conditioning the dynamics on the observed trajectory can represent multiple endpoints without the state sensitivity required by the predictable anchor in Proposition 11. However, tracking the audited trajectory may also absorb part of the anomaly. The relative importance of these effects depends on the fitted model and should be evaluated empirically.

One-class and reconstruction-based anomaly detection has an extensive literature. The closest work in robotics detects failed executions from models trained on normal trajectories, but generally localizes failures at the trajectory rather than step level, as noted by Yeh et al. (2026). The relevant statistical foundations include innovations, model-based fault detection and isolation, conformal inference, and multiple testing; Sections 2.2 and 5 summarize these connections.

\bigskip\begin{center}\large\textbf{Part I: Innovation auditing for predictable sequential models}\end{center}\medskip

\section{Setting}

\subsection{Trajectories, cost, and filtrations}

A trajectory is a sequence of operations $o_1, \ldots, o_T$ issued in response to a research question embedded as $x_0 \in \mathbb{R}^d$. Each operation carries an embedding $x_t = \phi(o_t) \in \mathbb{R}^d$ and an execution cost $c_t = \mathrm{cost}(o_t) > 0$, both functions of the same object. The run operates under a cost budget $C_{\max}$ fixed before execution, and every operation costs at least $c_{\min} > 0$, so

\[T \le T_{\max} := \lfloor C_{\max}/c_{\min}\rfloor\]

deterministically. This bound is used in Section 5.3 and motivates normalization by a fixed budget.

Two filtrations are required. Write

\[\mathcal{F}_t = \sigma(x_0, \ldots, x_t, c_1, \ldots, c_t), \qquad \mathcal{G}_t = \mathcal{F}_{t-1} \vee \sigma(c_t),\]

so $\mathcal{G}_t$ is what is known immediately before the embedding of operation $t$ is realized. Normalized time and its increment,

\[s_t = \frac{1}{C_{\max}}\sum_{i \le t} c_i, \qquad \Delta_t = \frac{c_t}{C_{\max}},\]

are $\mathcal{G}_t$-measurable but not $\mathcal{F}_{t-1}$-measurable, because the cost of operation $t$ is a property of that operation. All conditioning statements therefore use $\mathcal{G}_t$. Section 6.1 specifies the joint model required for this conditioning.

The termination time $T$ is a stopping time for $\{\mathcal{F}_t\}$, and $I \subseteq \{1, \ldots, T\}$ denotes the set of corrupted operations. Both are random. We require $I$ to be predictable in the sense that $\{t \in I\} \in \mathcal{G}_t$: whether operation $t$ violates the mechanism is determined by the state entering that operation, not by its realized outcome. The results do not condition on the realized values of $T$ or $I$. Conditioning on $T$ can bias earlier innovations when the stopping decision depends on the observed history.

\subsection{Predictive families and generalized residuals}

The model $\mathcal{M}$ is a predictable family of conditional laws: for each $t$, a probability measure $Q_t$ on $\mathbb{R}^d$ whose parameters are $\mathcal{G}_t$-measurable, together with a conditional law $P_t$ for $c_t$ given $\mathcal{F}_{t-1}$. Let $R_t$ denote the Rosenblatt transform of $Q_t$, that is, the vector of successive conditional distribution functions (Rosenblatt, 1952), and define the generalized residuals

\[u_t = R_t(x_t) \in [0,1]^d, \qquad \varrho_t = P_t\big((-\infty, c_t]\big) \in [0,1].\]

These are the generalized residuals of Cox and Snell (1968) in prequential form (Dawid, 1984). A score is any measurable $S : [0,1]^d \to \mathbb{R}$, and the audit statistic is $a_t = S(u_t)$.

The Gaussian bridge of Part II is the instance

\[Q_t = \mathcal{N}\big(\mu_t, \Sigma_t\big), \qquad \varepsilon_t = \Sigma_t^{-1/2}(x_t - \mu_t), \qquad u_t = \Phi(\varepsilon_t), \qquad S(u) = \|\Phi^{-1}(u)\|_2^2,\]

recovering $a_t = \|\varepsilon_t\|_2^2$. Part I does not depend on this Gaussian specification. The formulation in terms of predictive distributions also accommodates the multimodal predictive models required in Section 8.

For a correctly specified sequential model, whitened one-step prediction errors are innovations and therefore form a white sequence (Kailath, 1968). Quadratic forms in Kalman innovations are standard tools for model-based fault detection and isolation (Willsky, 1976; Basseville and Nikiforov, 1993), and the distinction in Section 3.2 between one-step and accumulated residuals follows the structured-residual literature (Gertler, 1998). The present contribution combines these residuals with distribution-free calibration and trajectory-level error control.

\subsection{Assumptions}

Each is invoked only where stated.

\textbf{(A0) Cost non-degeneracy.} $c_t \ge c_{\min} > 0$, and $\mathcal{L}(x_t \mid \mathcal{G}_t)$ is atomless. The second condition fails if cost determines the embedding, because conditioning on $\Delta_t$ would then condition on a function of $x_t$ and the predictive law of $x_t$ given $\mathcal{G}_t$ would be degenerate. The assumption is plausible when $\phi$ is lossy and cost depends on data volume as well as operation semantics, but it must be assessed for the deployed representation.

\textbf{(A1) Calibration exchangeability.} Calibration trajectories are i.i.d.\ draws from the population of valid trajectories, and the benign steps of a suspect trajectory are drawn from the same population.

\textbf{(A1$^\ast$) Calibration correctness.} The population law is $\mathcal{M}$. This assumption is strictly stronger than (A1) and is used only for pooling across steps within a trajectory and for the exact-null e-values in Section 5.4.

\textbf{(A2a) On-support benign validity.} For a benign step whose conditioning state lies in the training support, $\mathcal{L}(x_t \mid \mathcal{G}_t) = Q_t$.

\textbf{(A2b) Off-support benign conservatism.} For a benign step whose conditioning state lies outside the training support, the conditions of Theorem 7 hold.

\textbf{(A3) Predictable corruption.} $\{t \in I\} \in \mathcal{G}_t$.

\textbf{(A4) Regular drift.} In the bridge instance, $u_\theta(\cdot, s, \cdot)$ is $L$-Lipschitz in its first argument uniformly in $s$. Used in Sections 3.2 and 8 only.

\textbf{(A5) Uninformative selection.} The event that a trajectory is submitted for audit is conditionally independent of the benign generalized residuals given $x_0$ and the corrupted steps.

Assumptions (A2a) and (A2b) distinguish on-support and off-support behavior. Requiring benign steps to follow $\mathcal{M}$ exactly after arbitrary corrupted states would require the fitted model to be correct at states absent from training. Exact validity is therefore assumed only on support; off-support benign steps are required to satisfy the conservative-tail conditions in Section 7.

Assumption (A5) addresses selection into the audit sample. A prior process has already classified the trajectory as unsound, and that decision depends on the complete trajectory. Conditioning on selection can therefore alter the residual distribution. If selection is driven primarily by corrupted operations, (A5) may be a reasonable approximation. Section 5.2 treats settings in which selection also depends on benign residuals.

\section{Localization}

\subsection{Exact localization}

Corruption is modeled by replacing the predictive law: for $t \in I$, the conditional law of $x_t$ given $\mathcal{G}_t$ is $\tilde{Q}_t \ne Q_t$, with predictable parameters. An additive Gaussian mean shift is the special case $\tilde{Q}_t = \mathcal{N}(\mu_t + b_t, \Sigma_t)$.

\textbf{Theorem 1 (localization).} Assume (A0), (A2a), (A3), and that every $Q_t$ is atomless.

(i) For each $t$ and each bounded measurable $f$,
\[\mathbb{E}\big[f(u_t)\,\mathbf{1}\{t \le T,\ t \notin I\}\big] = \mathbb{E}[f(U)]\;\mathbb{P}(t \le T,\ t \notin I), \qquad U \sim \mathrm{Unif}[0,1]^d,\]
and for any $t_1 < \cdots < t_r$ the analogous product identity holds on the event that all of $t_1, \ldots, t_r$ are benign and at most $T$. The benign generalized residuals are i.i.d.\ uniform on the event that they exist, whatever corrupted states precede them.

(ii) For $t \in I$, conditionally on $\mathcal{G}_t$ the residual has law $R_{t\#}\tilde{Q}_t$ and the score has the corresponding pushforward law. No interaction between distinct corrupted steps arises.

The proof is given in Appendix A.1. Exactness follows from specifying the model at the level of realized discrete transitions. The events $\{t \le T\}$ and $\{t \notin I\}$ belong to $\mathcal{G}_t$, so the result accommodates both a stopping time and a random predictable corruption set. The identity is not conditioned on the realized value of $T$, because such conditioning can alter earlier innovation distributions.

Conditional on its input state, a correctly executed operation after a corruption follows the null predictive law. Under the stated assumptions, signal is therefore confined to $I$. Theorem 1 does not impose a causal ordering on the elements of $I$.

\textbf{Corollary 1 (Gaussian mean shift).} If $\tilde{Q}_t = \mathcal{N}(\mu_t + b_t, \Sigma_t)$ with $b_t$ being $\mathcal{G}_t$-measurable, then writing $\nu_t = \Sigma_t^{-1/2}b_t$ and $\lambda_t = \|\nu_t\|_2^2$, conditionally on $\mathcal{G}_t$ we have $a_t \sim \chi^2_d(\lambda_t)$, with tail probability strictly increasing in $\lambda_t$ (Johnson, Kotz and Balakrishnan, 1995).

\textbf{Corollary 2 (dispersion shifts and two-sided scoring).} If $\tilde{Q}_t = \mathcal{N}(\mu_t, \Sigma_t^{1/2}(I + \Gamma_t)\Sigma_t^{1/2})$ with $\Gamma_t \succ -I$ having eigenvalues $\gamma_{t,1}, \ldots, \gamma_{t,d}$, then $a_t \sim \sum_i (1 + \gamma_{t,i})\chi^2_1$. When $\Gamma_t \prec 0$ the score is stochastically smaller than $\chi^2_d$, and a one-sided upper-tail test has power below its level.

Corollary 2 identifies a class of failures that an upper-tail score cannot detect. Reduced exploration, including repeated templates or highly stereotyped analysis sequences, may produce residuals that are systematically smaller than the null distribution. A two-sided score addresses this alternative; Section 4.4 gives specific constructions.

\subsection{Multi-horizon residuals}

The natural competitor to one-step scoring reconstructs the trajectory from its start and scores each step by its distance to the rollout. In the bridge instance that comparison has an exact form. Consider the drift flow $\dot{z} = (\hat{x}_T - z)/(1-s) + u_\theta(z,s)$ on $[0,\bar{s}]$, $\bar{s} < 1$, and two solutions whose states at $s_0$ differ by $b \ne 0$.

\textbf{Theorem 2 (propagation envelope).} Under (A4), the gap $D(s)$ satisfies for all $s \in [s_0, \bar{s}]$

\[\|b\|\,\frac{1-s}{1-s_0}\,e^{-L(s-s_0)} \;\le\; \|D(s)\| \;\le\; \|b\|\,\frac{1-s}{1-s_0}\,e^{L(s-s_0)}.\]

The proof (Appendix A.2) uses a two-sided comparison argument (Grönwall, 1919). If the diffusion is state-independent, synchronous coupling cancels the noise and the envelope holds pathwise for the stochastic dynamics. With state-dependent diffusion, the result applies only to the drift flow.

The bounds imply decay toward the shared anchor only when $L < 1/(1-\bar{s})$, in which case both envelopes decrease. For larger $L$, the lower envelope still decreases, but the upper envelope increases. The result then establishes only a range of possible behavior. For example, at $L=20$, $s_0=0.5$, and $\bar{s}=0.9$, the envelope is $[6.7\times10^{-5},\,596]\|b\|$.

The comparison with the innovation residual relies on the strictly positive lower envelope. Reconstruction from $x_0$ assigns a nonzero deviation to every downstream operation, whereas a correctly specified one-step innovation is null after the corrupted transition. The two-sided bound does not determine whether the reconstruction deviation decreases, persists, or increases.

Free-running reconstruction from $x_0$ is only one member of a broader class of residuals. For $k \ge 1$, define the $k$-step predictive $Q_t^{(k)} = \mathcal{L}_\mathcal{M}(x_t \mid \mathcal{F}_{t-k}, c_{t-k+1}, \ldots, c_t)$ and the corresponding residual $u_t^{(k)}$ and score $a_t^{(k)}$. Theorem 1 does not apply verbatim to these residuals. Sliding windows overlap: $u_t^{(k)}$ and $u_{t+1}^{(k)}$ share $k-1$ transitions, and $u_t^{(k)}$ is not measurable with respect to the conditioning $\sigma$-field for $u_{t+1}^{(k)}$. The peeling argument in Appendix A.1 therefore does not yield a product identity. Marginal uniformity and independence across sufficiently separated indices remain available under the conditions below.

\textbf{Proposition 1 (width, validity, and dependence of the $k$-step channel).} Fix $k \ge 1$ and let $I = \{t_0\}$.

(i) \emph{Width.} $a_t^{(k)}$ has its null law for every $t \ge t_0 + k$, and its law is affected by the corruption only for $t_0 \le t \le t_0 + k - 1$.

(ii) \emph{Marginal validity.} If $\{t \le T\} \in \mathcal{F}_{t-k}$, so that the agent commits to the complete window at its start, then $u_t^{(k)}$ is marginally uniform on $\{t \le T,\ t \notin I\}$ by the argument of Appendix A.1 applied to the coarser filtration. Without this condition, halting decisions within the window can depend on its innovations, and marginal uniformity does not follow from the model.

(iii) \emph{Dependence.} For indices $t_1 < \cdots < t_r$ with $t_{j+1} - t_j \ge k$, the residuals $u_{t_1}^{(k)}, \ldots, u_{t_r}^{(k)}$ are jointly independent uniform on the corresponding benign event. For overlapping indices they are dependent, and no joint law is claimed.

For $t \ge t_0+k$, the conditioning $\sigma$-field contains $x_{t-k}$ with $t-k \ge t_0$, so the corrupted state is conditioned on rather than predicted through. Reconstruction from $x_0$ corresponds to $k=t$ and affects the complete suffix; the one-step innovation corresponds to $k=1$ and affects only the corrupted transition. Intermediate horizons yield intermediate localization widths.

Parts (ii) and (iii) determine which error-control results apply to the horizon channel. Benjamini--Yekutieli and e-BH require marginal super-uniformity and therefore accept the horizon channel. Proposition 6 requires independent null p-values and does not apply to overlapping $k$-step windows. It may be restricted to the one-step channel or evaluated on a non-overlapping subsequence $t \in \{t_0,t_0+k,t_0+2k,\ldots\}$, with a factor-$k$ reduction in the effective number of tests. If the window-commitment condition in part (ii) fails, the horizon channel can instead be calibrated under the exchangeability regime of Lemma 1(ii), stratified by trajectory length.

The wider localization window can increase power for temporally distributed perturbations. Let $\Sigma \equiv \sigma^2 I$ be state-independent and let the drift be locally the identity over a window of $k$ steps.

\textbf{Proposition 2 (accumulation gain).} Suppose a corruption of total displacement $b$ is injected evenly across $k$ consecutive steps, $b_t = b/k$. The one-step score at each of those steps has noncentrality $\|b\|^2/(k^2\sigma^2)$; the $k$-step score at the end of the window has noncentrality $\|b\|^2/(k\sigma^2)$.

Combining with the detection boundary of Proposition 3, the one-step channel detects such a corruption only when $\|b\|^2 \gtrsim k^2\sigma^2\sqrt{2d}$, while the $k$-step channel detects it when $\|b\|^2 \gtrsim k\sigma^2\sqrt{2d}$. In the intervening range

\[k\,\sigma^2\sqrt{2d} \;\lesssim\; \|b\|^2 \;<\; k^2\sigma^2\sqrt{2d}\]

the corruption is undetectable by one-step scoring but detectable by the $k$-step score. Thus, gradual perturbations are a limitation of the one-step horizon rather than of innovation scoring in general. A small set of horizons, such as $k \in \{1,4,16\}$, can recover these alternatives, with a corresponding increase in the number of hypotheses and localization width. Sequential accumulation methods derived from Page's CUSUM (Page, 1954) remain appropriate for detecting the onset of an unbounded regime change; Section 12 distinguishes that problem from bounded gradual perturbations.

\section{Choice of score}

Because every measurable functional of the generalized residual is pivotal under Theorem 1, the score affects power but not null validity. This section compares power across score choices.

\subsection{The detection boundary}

In the Gaussian instance with a mean shift, per-step detection reduces to testing $\chi^2_d$ against $\chi^2_d(\lambda)$.

\textbf{Proposition 3 (chi-square detection boundary).} Let $\chi^2_{d,1-\alpha}$ denote the $(1-\alpha)$-quantile of $\chi^2_d$ and $a \sim \chi^2_d(\lambda_d)$. As $d \to \infty$: if $\lambda_d/\sqrt{2d} \to 0$ then $\mathbb{P}(a > \chi^2_{d,1-\alpha}) \to \alpha$; if $\lambda_d/\sqrt{2d} \to c \in (0,\infty)$ then the power tends to $\Phi(c - z_\alpha)$; if $\lambda_d/\sqrt{2d} \to \infty$ the power tends to one.

The quadratic score has asymptotically negligible power when the noncentrality is $o(\sqrt{d})$. A perturbation confined to a few coordinates can therefore be diluted in a high-dimensional embedding.

\subsection{Sparse alternatives at finite dimension}

\textbf{Proposition 4 (asymptotic separation).} Suppose $\nu$ has $k_d$ nonzero coordinates of common magnitude $\delta_d = \sqrt{2(1+\eta)\log d}$ for some $\eta > 0$, with $k_d = o(\sqrt{d}/\log d)$. The level-$\alpha$ Bonferroni test rejecting when $\max_i |\varepsilon_i| > z_{\alpha/(2d)}$ has power tending to one, while the level-$\alpha$ chi-square test has power tending to $\alpha$.

The sparse-versus-dense dichotomy is classical in Gaussian sequence detection (Ingster and Suslina, 2003; Fan, 1996), and higher criticism interpolates between the regimes (Donoho and Jin, 2004).

Proposition 4 describes an asymptotic regime. The table below reports exact requirements at $\alpha=0.01$ and 80\% power: $\lambda^\ast$ is the noncentrality required by the chi-square channel, and $\lambda_k^\ast=k\delta^2$ is the requirement for the maximum channel when the signal occupies $k$ coordinates with common magnitude $\delta$.

\begin{center}\begin{tabular}{rrrrrrr}\hline
$d$ & $\chi^2_{d,0.99}$ & $\sqrt{2d}$ & $\lambda^\ast$ (dense) & $\lambda^\ast_{k=1}$ & $\lambda^\ast_{k=3}$ & $\lambda^\ast_{k=10}$ \\ \hline
8 & 20.1 & 4.0 & 20.6 & 16.5 & 27.2 & 47.7 \\
32 & 53.5 & 8.0 & 33.5 & 19.7 & 34.4 & 65.5 \\
128 & 168.1 & 16.0 & 59.0 & 22.9 & 41.7 & 84.3 \\
256 & 311.6 & 22.6 & 80.0 & 24.5 & 45.5 & 94.0 \\
\hline\end{tabular}\end{center}

At $d=256$, the maximum channel requires 3.3-fold less noncentrality than the chi-square channel for a one-coordinate perturbation and 1.8-fold less for a three-coordinate perturbation, but requires more for a ten-coordinate perturbation. The largest sparsity for which the maximum is more powerful is $k^\ast=1,2,5,7$ at $d=8,32,128,256$, respectively; at $d=256$, this is less than 3\% of coordinates. Thus, the finite-dimensional advantage is moderate and restricted to highly sparse alternatives. Both channels should be evaluated at the deployed dimension rather than selected from the asymptotic result alone.

Projection dimension affects power only when the projection preserves the perturbation. Suppose the corruption lies in a fixed subspace and the audit projects $\mathbb{R}^{256}$ onto $m$ dimensions. A signal-preserving projection lowers the required noncentrality from 80.0 at $m=256$ to 33.5 at $m=32$ and 20.6 at $m=8$. A projection selected without reference to the perturbation retains an expected fraction $m/d$ of the noncentrality; the required ambient noncentrality then increases as $m$ decreases (80 at $m=256$, 118 at $m=128$, 268 at $m=32$, and 660 at $m=8$). Dimension reduction therefore improves power only when it preserves the alternatives of interest. Section 11.2 gives a stronger reason to reduce effective dimension: the smallest attributable corruption has a floor of order $n^{-1/(d+2)}$, so dimension determines the rate at which calibration data improve attribution.

\subsection{Sparsity in the whitened basis}

Proposition 4 concerns sparsity of $\nu = \Sigma^{-1/2}b$, not of $b$.

\textbf{Proposition 5.} For general $\Sigma \succ 0$ there exist 1-sparse $b$ with $\Sigma^{-1/2}b$ having $d$ nonzero coordinates. If $\Sigma$ is diagonal, the supports of $b$ and $\Sigma^{-1/2}b$ coincide.

Whitened-coordinate sparsity is not a generic consequence of semantic localization; it depends on the covariance parameterization. Constraining $\Sigma_\theta$ to be diagonal preserves coordinate support. Under a diagonal-plus-low-rank form $\Lambda+UU^\top$, whitening mixes coordinates within the low-rank subspace, and the maximum channel should be evaluated in an appropriate covariance eigenbasis. The deployed covariance structure should be reported.

\subsection{Two-sided scores and channel combination}

Corollary 2 requires attention to the lower tail. Define $p_t^+$ from the upper tail of the calibration scores and $p_t^-$ from the lower, and use $2\min(p_t^+, p_t^-)$; or score directly by a functional sensitive to both, such as the Anderson--Darling distance of the coordinates of $u_t$ from uniformity, which is pivotal by Theorem 1 like any other.

Running several channels at level $q$ and taking their union generally inflates the error rate. A single combined statistic can instead be formed by standardizing each channel on the calibration set, taking the maximum, and conformalizing that maximum. Theorem 1 then gives a single pivotal statistic and one family of p-values for Section 5. Higher criticism (Donoho and Jin, 2004) provides another single statistic that adapts across sparsity regimes.

\subsection{Joint effect of the sensitivity parameters}

Sensitivity is governed jointly by the embedding dimension $d$, the variance floor $\sigma_{\min}^2$, and the off-support inflation rate $\gamma$ of Section 7. These are not independent choices. Write

\[\sigma^2_{\mathrm{eff}}(D) = \max\big\{\sigma^2_{\min},\ \sigma_0^2(1 + \gamma D^2)\big\}\]

for a step whose conditioning state lies at distance $D$ from the training support.

\textbf{Corollary 3 (minimum detectable perturbation).} At level $\alpha$ and power $1-\beta$, a mean-shift corruption $b$ is detectable by the chi-square channel only if
\[\|b\|_2^2 \;\ge\; \sigma^2_{\mathrm{eff}}(D)\,\sqrt{2d}\,\big(z_\alpha + z_{1-\beta}\big),\]
and by the maximum channel only if
\[\|b\|_\infty \;\ge\; \sigma_{\mathrm{eff}}(D)\,\big(z_{\alpha/(2d)} + z_{1-\beta}\big).\]

Increasing $d$ raises the dense threshold on $\|b\|$ at rate $d^{1/4}$ and the sparse threshold at rate $\sqrt{\log d}$. Increasing $\gamma$ or $\sigma_{\min}$ raises both thresholds linearly in $\sigma_{\mathrm{eff}}$. Robustness choices involving these parameters should therefore be reported together with their effects on detection sensitivity.

\section{Error control}

\subsection{Two calibration regimes}

Under (A1$^\ast$), all calibration residuals across all calibration trajectories are i.i.d.\ uniform, so their scores form an i.i.d.\ sample $\Omega = \{\omega_1, \ldots, \omega_n\}$ from the null score law. Pooling across steps and across trajectories is then exact by pivotality. For a suspect trajectory,

\[p_t = \frac{1 + \#\{j : \omega_j \ge a_t\}}{n+1}.\]

\textbf{Lemma 1 (validity).} (i) Under (A1$^\ast$), (A2a), (A3), and (A5), for each benign $t$, $\mathbb{P}(p_t \le u \mid t \le T,\ t \notin I) \le u$ for all $u \in [0,1]$ (Vovk, Gammerman and Shafer, 2005). (ii) Under (A1) and (A5) alone, the same result holds for a stratified p-value in which $\Omega$ is restricted to calibration scores at the same step index as $t$, or to the same bin of $s_{t-1}$, provided the benign test step and calibration trajectories are exchangeable within that stratum.

The two regimes impose different requirements on the fitted model. Pooling across steps within a trajectory requires correct specification, because pivotality makes scores at different positions comparable. Pooling across trajectories at a matched index requires only exchangeability and stable application of the fitted model, even under systematic misspecification. The second regime has lower resolution: with $B$ strata, the smallest attainable p-value is $1/(n/B+1)$. With many strata and only several hundred calibration trajectories, this discreteness can materially affect Benjamini--Yekutieli thresholds. A small number of prespecified bins of $s_{t-1}$ provides a practical compromise, and the binning rule should be reported.

\subsection{Selection of the audited trajectory}

The audit is applied after a prior process has classified a trajectory as unsound. The calibration results are valid under this conditioning only if (A5) holds. When selection depends mainly on corrupted operations, the distribution of benign residuals may be approximately preserved. When selection depends on the overall atypicality of the trajectory, benign residuals in selected trajectories can be stochastically larger than those in the calibration population, making the p-values anti-conservative.

Two approaches address selection. If the selection probability $\pi(\text{trajectory})$ can be estimated from reviewer scores or from labeled selected and unselected trajectories, weighted conformal prediction reweights the calibration distribution by the selected-to-unselected likelihood ratio (Tibshirani, Barber, Candès and Ramdas, 2019). Alternatively, calibration trajectories can be sampled from the selected population and independently adjudicated as valid. The latter avoids assumption (A5) but may yield a smaller calibration set. Section 5.7 develops a sensitivity analysis that does not require estimating the selection model.

\subsection{A p-value procedure with a fixed number of hypotheses}

Within one trajectory the null p-values share the calibration set and are therefore dependent even though the underlying null scores are independent. Corrupted-step scores may depend on earlier benign residuals through the state, since corruption parameters are predictable.

A second issue is that $T$ is random. The procedures of Benjamini and Hochberg (1995) and Benjamini and Yekutieli (2001) are formulated for a fixed number of hypotheses. Because $T \le T_{\max}$ deterministically, the family can be padded to $T_{\max}$ hypotheses by setting $p_t=1$ for $t>T$. The padded p-values are super-uniform and cannot be rejected, so the observed rejection set is unchanged and the number of hypotheses is fixed.

\textbf{Theorem 3 (arbitrary dependence).} Under (A1$^\ast$) or (A1) as appropriate, together with (A2a), (A3), (A5), applying the Benjamini--Yekutieli procedure to the padded family of $T_{\max}$ p-values at level $q/H_{T_{\max}}$, where $H_m = \sum_{k \le m} 1/k$, controls the trajectory false discovery rate at $q\,m_0/T_{\max} \le q$ under arbitrary joint dependence (Benjamini and Yekutieli, 2001).

A stronger result at the nominal Benjamini--Hochberg level requires additional dependence conditions. In the present corruption model, corrupted-step scores may depend on earlier benign residuals through predictable corruption parameters, so joint independence generally fails. It does hold in controlled fault-injection benchmarks when the corruption is fixed conditional on $x_0$. In that setting, the shared-calibration positive-dependence result of Bates, Candès, Lei, Romano and Sesia (2023) supports Benjamini--Hochberg at the nominal level. This result should therefore be stated as benchmark-specific rather than as a deployment-wide guarantee.

\subsection{An e-value procedure}

At $T_{\max}=40$, $H_{T_{\max}}$ is approximately 4.3, so the Benjamini--Yekutieli adjustment substantially reduces the rejection threshold. An e-value procedure avoids this harmonic factor.

Under (A1$^\ast$), the null score law is known exactly ($\chi^2_d$ in the Gaussian instance), so likelihood-ratio e-values are available in closed form. For a mixing distribution $\pi$ over noncentralities,

\[e_t \;=\; \int \frac{f_{d,\lambda}(a_t)}{f_d(a_t)}\,\pi(d\lambda), \qquad
\frac{f_{d,\lambda}(a)}{f_d(a)} \;=\; e^{-\lambda/2}\,\Gamma(d/2)\,\Big(\tfrac{\sqrt{\lambda a}}{2}\Big)^{-(d/2-1)} I_{d/2-1}\big(\sqrt{\lambda a}\big),\]

where $I_\nu$ is the modified Bessel function of the first kind. Each $e_t$ has null expectation one because it is a likelihood ratio. Under exchangeability alone, conformal p-values can be converted to e-values with any decreasing calibrator $f$ satisfying $\int_0^1 f=1$; for example, $f(p)=\kappa p^{\kappa-1}$ with $\kappa\in(0,1)$ (Vovk and Wang, 2021).

\textbf{Theorem 4 (e-BH).} Let $e_1, \ldots, e_{T_{\max}}$ be e-values for the padded family, valid for the nulls in the sense that $\mathbb{E}[e_t] \le 1$. Order them as $e_{(1)} \ge \cdots \ge e_{(T_{\max})}$ and set $k^\ast = \max\{k:e_{(k)} \ge T_{\max}/(qk)\}$. Rejecting the $k^\ast$ largest e-values controls the trajectory FDR at $q\,m_0/T_{\max} \le q$ under arbitrary dependence (Wang and Ramdas, 2022).

The absence of a harmonic factor does not imply uniformly greater power. A likelihood-ratio e-value requires a mixing distribution $\pi$, and power depends on its mass near the relevant alternatives; Corollary 3 identifies the corresponding detection scale. Converting conformal p-values to e-values also reduces efficiency. Relative power should therefore be evaluated for the intended trajectory length and alternative class. Boosting against a known null can recover part of this loss (Wang and Ramdas, 2022).

\subsection{Per-report false discovery proportion}

Trajectory-level false discovery rate is an expectation over an ensemble of audited trajectories. It is not a high-probability guarantee for the realized false discovery proportion in a single report.

A high-probability statement for a single report requires a simultaneous bound on the realized false discovery proportion. Under (A1$^\ast$) and (A2a), the benign p-values from the one-step channel are i.i.d.\ uniform, so the Dvoretzky--Kiefer--Wolfowitz inequality with Massart's constant applies (Dvoretzky, Kiefer and Wolfowitz, 1956; Massart, 1990). Independence is required: overlapping horizon scores in Section 3.2 and the anchor correction in Section 6.3 do not satisfy this condition.

\textbf{Proposition 6 (simultaneous FDP bound, one-step channel).} With probability at least $1-\delta$, simultaneously over all thresholds $s \in [0,1]$,
\[\mathrm{FDP}(s) \;\le\; \frac{T s + \sqrt{T\log(2/\delta)/2}}{1 \vee R(s)},\]
where $R(s)$ is the number of p-values at most $s$. Sharper simultaneous bounds tuned to BH-type threshold families are available (Katsevich and Ramdas, 2020).

At $\delta=0.1$, the additive term equals 3.9 flags at $T=10$, 5.5 at $T=20$, and 8.7 at $T=50$. The bound is therefore uninformative unless the number of reported operations is large relative to typical trajectory length. At the trajectory lengths considered here, Sections 5.3 and 5.4 provide long-run FDR guarantees rather than report-specific probability statements. The trajectory remains the appropriate unit of multiplicity adjustment because the number of true errors may vary substantially across analyses, but the guarantee is an ensemble property.

\subsection{Robustness to misspecification}

\textbf{Theorem 5 (total-variation robustness).} Suppose corrupted steps have arbitrary predictable kernels, and each benign step's true conditional law given $\mathcal{G}_t$ is within total variation $\delta_t$ of $Q_t$, uniformly over histories. Let $\psi$ be any flagging procedure, measurable in the scores and the calibration set, whose FDR is at most $q$ when benign steps follow $\mathcal{M}$. Then under the true dynamics $\mathrm{FDR}(\psi) \le q + \sum_{t \notin I}\delta_t$.

The proof (Appendix A.7) is a sequential maximal coupling (Lindvall, 2002).

Applying the same coupling argument to calibration misspecification yields an additive term $n\epsilon$ for $n$ calibration points whose laws are each within $\epsilon$ of the model. This bound is generally uninformative for realistic $n$. Validity depends on the calibration empirical distribution function, which permits a sharper concentration argument.

\textbf{Proposition 7 (sharpened calibration term).} Suppose each calibration score is drawn independently from a law $G$ with $\|G - F\|_\infty \le \epsilon$, where $F$ is the model null score law. Fix $\delta \in (0,1)$ and set
\[\eta \;=\; \epsilon + \sqrt{\frac{\log(2/\delta)}{2n}} + \frac{1}{n+1}.\]
Then with probability at least $1 - \delta$ the inflated p-values $\tilde{p}_t = \min(1, p_t + \eta)$ are super-uniform for benign $t$, so every guarantee of Sections 5.3 and 5.4 holds for $\tilde{p}$ with an additional $\delta$ added to the FDR.

The correction improves the bound from $n\epsilon$ to $\epsilon+O(n^{-1/2})$ and is directly implementable by inflating each p-value by $\eta$. Estimation error requires no separate term because assumptions (A1), (A2a), and (A2b) concern the deployed kernels; discrepancies between the fitted model and the data-generating law are included in these assumptions.

A remaining limitation is that total variation between transition kernels can be much larger than total variation between the induced score laws, although flagging depends on the trajectory only through the scores. A bound stated in terms of score laws would be sharper and estimable from held-out valid trajectories. Such a bound is not derived here because the sequential coupling requires equality of states, whereas equality of scores does not imply equality of states. Section 9 describes diagnostics that can be evaluated directly.

\subsection{Sensitivity to the selection mechanism}

The submission propensity is difficult to model because the dependence of the upstream judgment on benign operations is not directly observable. A sensitivity analysis can instead bound the propensity variation and report how much selection would be required to remove each flag.

\textbf{Theorem 6 (selection sensitivity).} Let $S$ be the event that a trajectory is submitted for audit, $\pi(\tau) = \mathbb{P}(S \mid \tau)$ its propensity, and suppose that across valid trajectories the propensity odds vary by at most a factor $\Gamma \ge 1$:
\[\Gamma^{-1} \;\le\; \frac{\pi(\tau)\big/\big(1-\pi(\tau)\big)}{\pi(\tau')\big/\big(1-\pi(\tau')\big)} \;\le\; \Gamma \qquad \text{for all valid } \tau, \tau'.\]
Then for every benign step, $\mathbb{P}(p_t \le u \mid S) \le \Gamma u$ for all $u \in [0,1]$, and Benjamini--Yekutieli or e-BH applied at level $q/\Gamma$ controls the trajectory false discovery rate at $q$ within the selected population.

The proof (Appendix A.11) follows because the odds bound implies $\pi(\tau) \le \Gamma\pi(\tau')$ pointwise, so selection can increase any benign-event probability by at most $\Gamma$. When $\Gamma=1$, submission is exchangeable and (A5) holds. A value $\Gamma=2$ permits a twofold difference in submission probabilities and requires a corresponding twofold reduction in the operating level.

For a Benjamini--Yekutieli rejection at rank $k$ with p-value $p_{(k)}$, the largest propensity ratio under which the rejection remains is
\[\Gamma^\ast \;=\; \frac{q\,k}{T_{\max}\,H_{T_{\max}}\;p_{(k)}}.\]
Thus, each flag can be accompanied by the selection-strength value that would remove it. Larger values of $\Gamma^\ast$ indicate greater robustness to selection; values near one indicate substantial sensitivity. The quantity is available from the observed p-values and requires no fitted selection model.

\bigskip\begin{center}\large\textbf{Part II: The bridge instantiation}\end{center}\medskip

\section{The bridge model}

\subsection{Time, cost, and predictability of the grid}

Normalization by realized total cost, $\sum_{i\le T}c_i$, makes early time points depend on future operation costs and is therefore incompatible with predictable conditioning. Budget normalization, $s_t=C_{\max}^{-1}\sum_{i\le t}c_i$, removes this dependence. It also ensures $s_T<1$ on valid runs, so the bridge drift does not reach the $1/(1-s)$ singularity, and gives the deterministic bound $T\le T_{\max}$ used in Section 5.3.

Budget normalization alone does not make the grid predictable. The increment $\Delta_t=c_t/C_{\max}$ and the embedding $x_t$ are both properties of operation $t$. A model that conditions the law of $x_t$ on $\Delta_t$ therefore conditions on another feature of the same operation. Under (A0), this conditional law is well defined and non-degenerate, but the joint distribution must be factorized explicitly:

\[p(c_t, x_t \mid \mathcal{F}_{t-1}) \;=\; \underbrace{p_\psi(c_t \mid \mathcal{F}_{t-1})}_{\text{cost model}} \;\cdot\; \underbrace{p_\theta(x_t \mid c_t, \mathcal{F}_{t-1})}_{\text{embedding model, } = Q_t}.\]

The second factor is the Gaussian bridge transition defined below and is conditioned on $\mathcal{G}_t$. The first factor is necessary unless operation cost is treated as exogenous.

Execution cost provides an additional audit channel. An operation that processes substantially fewer records than predicted may indicate an unintended cohort restriction, whereas a substantially larger cost may indicate a join or keying error. The cost residual $\varrho_t$ can be standardized on the calibration set, combined with the embedding score before conformalization, and represented by one p-value per operation for the multiplicity procedures in Section 5.

A deployment that does not model cost can set $\Delta_t\equiv 1/T_{\max}$ and include realized cost only as a covariate of $\Sigma_\theta$ through $\mathcal{F}_{t-1}$. This alternative is predictable but does not preserve the geometry of cost-normalized time.

\subsection{Dynamics}

Let $\hat{x}_T^{(t)} = g_\eta(x_0, x_{t-1}, s_{t-1})$ be a terminal anchor. For $t = 1, \ldots, T$,

\[x_t = \mu_t(x_{t-1}) + \Sigma_t(x_{t-1})^{1/2}\xi_t, \qquad \xi_t \sim \mathcal{N}(0, I_d)\ \text{i.i.d.},\]
\[\mu_t(x) = x + \left[\frac{\hat{x}_T^{(t)} - x}{1 - s_{t-1}} + u_\theta\big(x, s_{t-1}, \hat{x}_T^{(t)}\big)\right]\Delta_t, \qquad \Sigma_t(x) = \Sigma_\theta\big(x, s_{t-1}, \Delta_t\big) \succeq \sigma_{\min}^2 I_d.\]

Three aspects of this specification require comment. First, the anchor is re-estimated from the current state at each step rather than fixed at $g_\eta(x_0)$. Because $\hat{x}_T^{(t)}$ is $\mathcal{G}_t$-measurable, the conditioning statements in Part I remain valid. The anchor head should be trained against held-out realized endpoints rather than through the trajectory likelihood; otherwise, the degenerate solution $\hat{x}_T^{(t)}\to x_{t-1}$ can eliminate the bridge drift while increasing the likelihood.

Second, the covariance is a free function of state and time rather than being proportional to $\Delta_t$. Proportionality would impose diffusive $\sqrt{\Delta_t}$ scaling, which can underweight operations that have low execution cost but large semantic effects, such as a restrictive cohort filter. Section 4.3 further motivates a diagonal or diagonal-plus-low-rank covariance so that sparse alternatives remain interpretable after whitening.

Third, the covariance floor prevents degeneracy of the heteroscedastic Gaussian likelihood when fitted variances collapse on interpolated observations (Seitzer et al., 2022). Corollary 3 also shows that the floor determines detection sensitivity and should be selected with reference to a minimum perturbation of scientific interest.

The mean is an Euler discretization (Kloeden and Platen, 1992) of a bridge stochastic differential equation whose leading drift is the $h$-transform of Brownian motion pinned at the anchor (Doob, 1957; Rogers and Williams, 2000). The discrete recursion, rather than the continuous-time equation, is the statistical model. Its transition law is therefore exact by construction, whereas an exact transition density for the nonlinear stochastic differential equation is not assumed.

\subsection{The anchor as a nuisance parameter}

Anchor error affects every step through a known loading structure. If the fitted anchor is $\hat{x}_T^{(t)}$ but the trajectory is directed toward $\hat{x}_T^{(t)}+g$, the true conditional mean differs from $\mu_t$ by $g\,w_t$, where

\[w_t = \frac{\Delta_t}{1 - s_{t-1}}\]

is known. In the isotropic case $\Sigma_t=\sigma_t^2 I$, define $v_t=w_t/\sigma_t$ and $W=\sum_{t\le T}v_t^2$. Then $\varepsilon_t=\xi_t+v_t g$, which is a fixed-effect model with a vector parameter $g$ and known scalar loadings.

\textbf{Proposition 8 (anchor projection and induced contamination).} For $\varrho \ge 0$ let $\hat{g}_\varrho = (W+\varrho)^{-1}\sum_t v_t\varepsilon_t$, $\tilde{\varepsilon}_t = \varepsilon_t - v_t\hat{g}_\varrho$ and $h_t = v_t^2/(W+\varrho)$, with $\sum_t h_t = W/(W+\varrho) \le 1$.

(i) If no step is corrupted then $\tilde{a}_t = \|\tilde{\varepsilon}_t\|^2/(1-h_t)$ is marginally $\chi^2_d$ for every $g$, but the corrected residuals are cross-correlated: $\mathrm{Cov}(\tilde{\varepsilon}_t, \tilde{\varepsilon}_s) = -v_t v_s(W+\varrho)^{-1}I_d$ for $s \ne t$.

(ii) A mean shift at a corrupted step $t$ contributes noncentrality $(1-h_t)\lambda_t$ at $t$ and $\mu_s = h_s h_t\lambda_t/(1-h_s)$ at every benign $s \ne t$.

Part (ii) concerns validity rather than only power. Because the anchor estimate uses all operations, a corruption at one step induces a nonzero mean at otherwise benign steps, so their null hypotheses in Theorems 3 and 4 no longer hold. The effect is largest near the end of the trajectory, where $w_t$ and leverage are greatest. For example, when $h_s=h_t=0.3$ and $\lambda_t=100$, the induced noncentrality at step $s$ is 12.9.

The induced misspecification can be bounded. Since the total variation distance between $\mathcal{N}(\theta,I_d)$ and $\mathcal{N}(0,I_d)$ is at most $\|\theta\|/\sqrt{2\pi}$, Theorem 5 gives

\[\sum_{s \ne t}\delta_s \;\le\; \frac{\sqrt{h_t\lambda_t}}{\sqrt{2\pi}}\sum_{s \ne t}\sqrt{\frac{h_s}{1-h_s}},\]

which grows as $\sqrt{\lambda_t}$ and is unbounded in the corruption magnitude. Shrinkage through $\varrho$ changes the scale but does not remove this dependence.

A bounded-influence estimator addresses this problem. Estimating $g$ from $\{\varepsilon_t/v_t\}$ with a Huberized or trimmed procedure caps the influence of any one operation independently of $\lambda_t$, so the induced means and total-variation bound remain uniformly bounded. The corrected score then need not have an exact $\chi^2_d$ null distribution. Applying the same correction to calibration trajectories and using the exchangeability regime of Lemma 1(ii) avoids reliance on that parametric null.

Part (i) has two additional consequences. The induced cross-correlation invalidates Proposition 6, which requires independent nulls, but does not affect Benjamini--Yekutieli or e-BH\@. For non-isotropic $\Sigma_t$, the same construction uses $W_t=w_t\Sigma_t^{-1/2}$ and leverage matrices $H_t$, with corrected residual covariance $I-H_t$.

Anchor correction is therefore most appropriate as a sensitivity analysis reported alongside the uncorrected scores and per-step leverage, rather than as the primary score. Because leverage is greatest near the terminal operations, the correction further reduces sensitivity in the region already affected by a point-valued terminal predictive.

\section{Off-support behavior}

\subsection{Sufficient conditions for off-support conservatism}

A learned heteroscedastic variance does not, by itself, control false alarms on out-of-distribution trajectories; an unconstrained variance head may extrapolate to arbitrarily small values. Sufficient conditions require variance to increase with distance from the training support and require the mean extrapolation error to be bounded by that distance. Spectral normalization can provide the required Lipschitz control (Miyato et al., 2018; Liu et al., 2020). Under a bi-Lipschitz representation, feature-space distance to the training set also controls input-space distance up to fixed constants.

For composition with the procedures in Section 5, the relevant condition is upper-tail dominance rather than validity at a single threshold.

\textbf{Theorem 7 (off-support tail conservatism).} Consider a benign step whose conditioning state lies at distance $D \ge d_0$ from the training support. Suppose the model is isotropic with variance $\sigma^2(D) = \sigma_0^2(1 + \gamma D^2)$; the true conditional law has mean within $\kappa D + \eta_0$ of the model mean and covariance $\preceq \bar\sigma^2 I_d$. Put
\[A_\infty = \frac{2(\kappa + \eta_0/d_0)^2}{\sigma_0^2\,\gamma}, \qquad B_0 = \frac{2\bar\sigma^2}{\sigma_0^2(1 + \gamma d_0^2)}.\]
If $B_0 < 1$ then for every $u \ge A_\infty/(1 - B_0)$,
\[\mathbb{P}(a_t > u) \;\le\; \mathbb{P}(\chi^2_d > u).\]
In particular, if $A_\infty/(1-B_0) \le \chi^2_{d,1-\alpha}$, the benign p-value is $\alpha$-tail super-uniform: $\mathbb{P}(p_t \le u) \le u$ for every $u \le \alpha$.

Both $A_\infty$ and $B_0$ decrease with $\gamma$, so the conditions hold above an explicit inflation threshold. With a spectrally normalized mean head of Lipschitz constant $L_\mu$ and a true mean map assumed to be $L^\ast$-Lipschitz, $\kappa\le L_\mu+L^\ast$, while $\eta_0$ is the on-support fit error. The assumption of Gaussian off-support deviations can be replaced by a sub-Gaussian condition, with corresponding changes to the constants.

Benjamini--Hochberg, Benjamini--Yekutieli, and e-BH use super-uniformity only at levels below their operating level. Thus, when $q\le\alpha$, Theorem 7 supplies assumption (A2b) and extends the error-control results of Part I to trajectories that leave the training support.

\textbf{Proposition 9 (off-support detection threshold).} At a corrupted step at the same distance $D$, a shift $b$ yields effective noncentrality $\lambda_{\mathrm{eff}} = \|b\|^2/\sigma^2(D)$. By Corollary 3, detection at level $\alpha$ and power $1-\beta$ requires $\|b\|^2 \ge \sigma_0^2(1 + \gamma D^2)\sqrt{2d}\,(z_\alpha + z_{1-\beta})$.

The variance inflation that makes off-support null scores conservative reduces the noncentrality of genuine off-support perturbations by the same factor.

\subsection{Inherited corruption}

Exact on-path validity at every benign step is incompatible with unconstrained off-support behavior after a corruption. A downstream benign operation may receive a conditioning state that is absent from the training support even though its transition mechanism is valid. Theorem 7 addresses this case through tail conservatism rather than exact equality.

Theorem 7 yields the following conservative localization result.

\textbf{Corollary 4.} Let $t \notin I$ follow a corrupted step, so that its conditioning state lies at distance $D \ge d_0$ from the training support, and suppose the conditions of Theorem 7 hold. Then $\mathbb{P}(a_t > u) \le \mathbb{P}(\chi^2_d > u)$ for all $u \ge A_\infty/(1-B_0)$, and the step is flagged with probability at most its nominal level.

Exact localization is replaced by conservative localization. Downstream residuals are exactly null when the model is correct at the inherited state and have upper tails no heavier than the null when variance inflation is required off support.

The same variance inflation also reduces sensitivity to a second genuine error committed after the trajectory has left the training support, by the factor in Proposition 9. Control of downstream false alarms and detection of subsequent errors therefore constitute a direct trade-off.

\section{Terminal anchoring with multimodal endpoints}

Valid analyses of the same question may terminate at different endpoints, such as distinct defensible covariate sets or compatible estimands. A point-valued anchor conditions the drift on one endpoint. Under (A4), routing trajectories to well-separated modes late in the run requires state sensitivity that grows as $1/(1-s)$, which a uniformly Lipschitz $u_\theta$ cannot provide. In addition, an $\ell_2$-trained point anchor converges to the conditional mean of the valid endpoints, which may lie in a low-density region between modes.

\textbf{Proposition 10 (terminal multimodality).} Consider the terminal step with isotropic model variance $\sigma_T^2$, flag threshold $\tau$, and (A4) in force. Let $z, z'$ be two benign pre-terminal states with $\|z - z'\| \le w$ whose true terminal laws are $\mathcal{N}(m_i, \sigma_c^2 I_d)$ with $\|m_1 - m_2\| \ge 2r$. Let $L_g$ be the Lipschitz constant of the anchor head in its state argument, put $h_T = 1 - s_{T-1}$ and $\kappa_T = 1 + L_g\Delta_T/h_T + L\Delta_T$, and set $\rho = r - w\kappa_T/2$. If $\rho \ge 2\sigma_c(\sqrt{d} + t)$, then at least one of the following holds: (a) the benign terminal flag probability at one of the two states exceeds $1 - e^{-t^2/2}$; or (b) $\sigma_T^2 \ge \rho^2/(4\tau)$, in which case every terminal shift with $\|b\|^2 \le c_1\rho^2/\sqrt{d}$ for a universal constant $c_1$ is undetectable at the boundary of Proposition 3.

Maximum-likelihood training strongly penalizes under-coverage and may therefore favor alternative (b), with an inflated terminal variance. Under this behavior, endpoint multimodality appears as reduced sensitivity to corruptions in the final operations rather than as an increased terminal false-positive rate. Section 9 describes an empirical test of this prediction.

Two modifications are compatible with the framework in Part I.

The predictable anchor in Section 6.2 is useful when the terminal mode is identifiable from the pre-terminal state. The function $g_\eta(x_0,x_{t-1},s_{t-1})$ can then select an endpoint without increasing the Lipschitz constant of the complete drift. Appendix A.10 shows that the dichotomy applies only while $\rho>0$ and that this condition fails when the anchor head's state sensitivity exceeds $(2r/w-1-L\Delta_T)h_T/\Delta_T$. Exceeding this threshold provides capacity to separate the modes but does not establish accurate routing. Section 8.1 gives an additional accuracy condition. A point anchor remains inadequate when more than one endpoint is plausible from the same pre-terminal state.

A more general modification replaces the point anchor with a multimodal predictive distribution, such as a mixture over endpoints or a conditional normalizing flow. Because Part I is formulated through generalized residuals for arbitrary predictive families, this modification changes $Q_t$ but leaves Theorem 1, Lemma 1, Theorems 3--5, and Proposition 6 unchanged. The Gaussian closed forms in Section 4 and the bridge-specific results in Part II would require corresponding modification.

\subsection{Routing capacity and off-support sensitivity}

The anchor head contributes to the terminal mean and therefore to the extrapolation constant in Theorem 7. With $\kappa_T=1+L_g\Delta_T/h_T+L\Delta_T$, the terminal mean map satisfies $\kappa\le\kappa_T+L^\ast$.

\textbf{Proposition 11 (off-support consequence of routing capacity).} Under the conditions of Theorem 7 applied at the terminal step, the second design condition forces
\[\gamma \;\ge\; \frac{2\big(\kappa_T + L^\ast + \eta_0/d_0\big)^2}{\sigma_0^2\,\chi^2_{d,1-\alpha}\,(1-B_0)},\]
and hence, by Proposition 9 and for $\gamma D^2 \gg 1$, a minimum detectable terminal perturbation at distance $D$ of
\[\|b\|^2_{\min} \;\ge\; \frac{2\big(\kappa_T + L^\ast + \eta_0/d_0\big)^2\,D^2\,\sqrt{2d}\,(z_\alpha + z_{1-\beta})}{\chi^2_{d,1-\alpha}\,(1-B_0)},\]
in which $\sigma_0^2$ cancels. If $L_g$ exceeds the separation threshold in Proposition 10, then $\kappa_T>2r/w$, so the off-support minimum detectable perturbation grows at least quadratically with the mode-separation ratio $r/w$.

Increasing the anchor's capacity to separate endpoints also increases the Lipschitz constant of the terminal mean. Theorem 7 then requires greater variance inflation, and Proposition 9 implies lower off-support sensitivity. The effect is quadratic in the separation ratio. This trade-off is less restrictive near the training support, where $D$ is small, but can be substantial for trajectories that are both strongly routed and far off support.

Exceeding the capacity threshold removes the obstruction in Proposition 10 but does not ensure that the fitted anchor maps each state to the appropriate mode. An accuracy condition is also required.

\textbf{Proposition 12 (accurate routing avoids both failure modes).} In the setting of Proposition 10, suppose the anchor routes accurately in the sense that $\|\mu(z) - m(z)\| \le e_0$ at every benign pre-terminal state $z$, where $m(z)$ is the mean of that state's true terminal law. If
\[\sigma_T^2 \;\ge\; 2\sigma_c^2 + \frac{2e_0^2}{\chi^2_{d,1-\alpha}},\]
then the benign terminal score is tail-dominated by $\chi^2_d$ above $\chi^2_{d,1-\alpha}$, so alternative (a) does not occur. A terminal shift is detectable at level $\alpha$ and power $1-\beta$ once $\|b\|^2 \ge \sigma_T^2\sqrt{2d}(z_\alpha + z_{1-\beta})$, which is of order $\sigma_c^2\sqrt{2d}$ when $e_0^2 \lesssim \sigma_c^2\chi^2_{d,1-\alpha}$.

Under alternative (b) of Proposition 10, blindness extends to $\|b\|^2\lesssim\rho^2/\sqrt{d}$. Under accurate routing, it extends only to order $\sigma_c^2\sqrt{d}$. The improvement factor is $\rho^2/(\sigma_c^2d)$, which is at least four under the hypothesis $\rho\ge2\sigma_c(\sqrt{d}+t)$ and increases with mode separation. Accurate routing therefore replaces the between-mode separation by the within-mode spread and routing error as the scale governing terminal sensitivity. It cannot remove the intrinsic detection boundary associated with terminal spread $\sigma_c$.

The routing modification therefore requires three conditions: sufficient capacity to separate the modes (Proposition 10), sufficient accuracy to route each state to its mode (Proposition 12), and acceptable off-support sensitivity at the resulting Lipschitz constant (Proposition 11). A mixture predictive can represent separated modes without requiring a steep point-valued anchor and may therefore avoid the first and third constraints.

\section{Diagnostics}

Two assumptions can be assessed using data already available in deployment.

Under (A1$^\ast$), pooled calibration scores follow $\chi^2_d$. Quantile--quantile plots and Kolmogorov--Smirnov or Anderson--Darling statistics on held-out valid trajectories can assess this implication. Diagnostics should be reported both overall and within strata of $s_{t-1}$, because localized misspecification may be obscured by pooling. Rejection indicates that the uninflated guarantees in Section 5.3 do not apply; failure to reject does not establish correct specification. Deployment documentation should state this asymmetry.

Four simulation studies can assess whether the fitted implementation exhibits the predicted behavior. First, trajectories sampled from $\mathcal{M}$ should produce $\chi^2_d$ benign scores and null downstream scores after an injected shift. Second, power should be evaluated across noncentralities at the deployed dimension to assess the approximation in Proposition 3. Third, one-step, intermediate-horizon, and free-running scores can be compared on the same injected perturbation to measure the localization window in Proposition 1. Fourth, a bimodal-endpoint simulation can evaluate terminal sensitivity.

The final simulation directly tests the prediction of Proposition 10. If a model with demonstrably multimodal endpoints exhibits elevated terminal false-positive rates rather than reduced terminal sensitivity, then training has selected alternative (a) rather than alternative (b), and the assumed training behavior in Section 8 does not hold for that implementation.

\bigskip\begin{center}\large\textbf{Part III: Population-level auditing and identifiability}\end{center}\medskip

\section{Recurrent errors across trajectories}

The preceding results concern one trajectory. In deployment, recurrent execution errors may appear across many trajectories, including repeated cohort restrictions, join failures, or omitted adjustments. Pooling repeated occurrences provides information that is unavailable in a single-trajectory analysis.

Let each operation have a type label $c(o_t) \in \{1,\ldots,C\}$, defined by the invoked tool or by a cluster of operation embeddings. Across $M$ audited trajectories, let $\mathcal{T}_c$ contain the steps of type $c$, with $m_c=|\mathcal{T}_c|$, and let $H_c$ denote the hypothesis that type $c$ is executed validly wherever it occurs. This estimand concerns recurrent type-level departures rather than the corrupted set within one trajectory.

\textbf{Proposition 13 (recurrence gain).} Suppose (A1$^\ast$) and (A2a) hold, a type-$c$ defect contributes noncentrality $\lambda$ at each of its occurrences, and the occurrences lie in distinct trajectories. Then the pooled score $A_c = \sum_{t \in \mathcal{T}_c} a_t$ is $\chi^2_{m_c d}(m_c\lambda)$, and detection at level $\alpha$ with power $1-\beta$ requires
\[\lambda \;\ge\; \sqrt{\frac{2d}{m_c}}\,\big(z_\alpha + z_{1-\beta}\big),\]
a factor $\sqrt{m_c}$ below the per-step boundary of Proposition 3.

For type-level error control, averaging the e-values in Section 5.4 within a type yields a valid type-level e-value. Applying e-BH across the $C$ types controls type-level FDR at $q$ under arbitrary dependence, including dependence among occurrences within the same trajectory.

The recurrence gain applies only to the noise-limited component of the detection boundary.

\textbf{Corollary 5 (recurrence does not resolve confusability).} The $\sqrt{m_c}$ improvement applies to the noise-limited boundary of Section 11.2 only. The confusability floor of Theorem 9 concerns the distinction between a corruption and an unmodelled feature of the \emph{valid} dynamics, and is governed by the number of valid calibration trajectories $n$, never by the number of audited occurrences $m_c$.

Audited occurrences cannot resolve uncertainty about the valid mechanism itself. A recurrent departure concentrated in one region of state space may be observationally similar to systematic misspecification of the valid drift. Recurrence increases statistical power against known null dynamics but does not reduce the identifiability limit. Analyses that pool across trajectories should distinguish these two sources of uncertainty.

\section{Identifiability and detection limits}

The guarantees in Parts I and II address the error rate of the reported set. This section addresses the interpretation of that set. Three results limit the departures that can be identified from observational trajectories.

\subsection{Testability of the estimand}

Exact inequality between transition kernels is not a statistically separated estimand.

\textbf{Theorem 8 (non-testability of the exact-departure estimand).} Fix a valid mechanism $K^\ast$ with transition densities, a step $t_0 < T$, and $\eta \in (0,1)$. There exist mechanisms $K^{(1)}, K^{(2)}$ with
\[I(K^{(1)}) = \{t_0\}, \qquad I(K^{(2)}) = \{t_0, t_0+1, \ldots, T\}, \qquad \mathrm{TV}\big(P^{(1)}, P^{(2)}\big) \le \eta,\]
where $I(K) = \{t \le T : K_t \ne K^\ast_t\}$ and $P^{(i)}$ is the law of the audited trajectory under $K^{(i)}$, the calibration law being the same under both. Consequently every auditor $\psi$ satisfies
\[\mathbb{E}_{1}\big|\psi \,\triangle\, I(K^{(1)})\big| + \mathbb{E}_{2}\big|\psi \,\triangle\, I(K^{(2)})\big| \;\ge\; (T - t_0)(1 - \eta).\]

The construction in Appendix A.13 modifies every kernel after $t_0$ by an $\epsilon$-mixture of the valid kernel and an alternative. Each subsequent kernel differs from $K^\ast$, while the total variation distance between the resulting trajectory laws can be made arbitrarily small. Thus, a set estimator that is correct under one mechanism can have large symmetric-difference loss under an observationally indistinguishable mechanism.

\textbf{Corollary 6.} No auditor estimates $I(K)$ consistently, at any sample size, for any score, under any assumption on $K^\ast$ that does not restrict the size of departures.

The estimand must therefore include a minimum departure magnitude. For $\delta>0$, define $I_\delta(K)=\{t:\mathrm{TV}(K_t,K_t^\ast)>\delta\}$ and leave departures below $\delta$ in an indifference region. The audit then reports operations whose conditional mechanisms differ from the valid mechanism by more than a prespecified threshold. Section 2.3 should be interpreted with this qualification.

Assumption (A2a) provides a positive result under competence preservation.

\textbf{Proposition 14 (factorization under competence).} Suppose $K^\ast$ is known and every benign step satisfies $K_t = K^\ast_t$ conditionally on $\mathcal{G}_t$ whatever the past states. Then the likelihood ratio of any corruption hypothesis against the null factorizes over steps, the generalized residuals are jointly sufficient for $I_\delta$, and every admissible auditor is a function of them. If the mechanism may instead change after an error and remain changed, no such factorization holds: the minimal sufficient statistic is the whole trajectory, the natural hypothesis family is nested rather than indexed by individual steps, and step-wise testing is inadmissible.

Competence preservation is the condition under which step-level attribution is well posed. If the mechanism changes after an error and remains changed, the appropriate estimand is the time at which the process leaves the valid regime, not a set of independent corrupted operations. Controlled replay and counterfactual intervention can restore step-level identification by fixing the downstream mechanism.

\subsection{A detection floor from finite calibration}

After introducing a magnitude threshold, the attribution rate is limited by uncertainty in the valid dynamics. A local mean shift at one operation can be observationally equivalent to a legitimate local feature of the valid drift. Distinguishing these cases requires sufficient calibration trajectories in the same region of state space and is therefore a local nonparametric estimation problem.

Fix a class $\mathcal{U}(L)$ of drifts $L$-Lipschitz in the state, noise level $\sigma^2$, per-step increment $\Delta$, and a visited-state density bounded above by $c$. Write $N = nT$ for the total number of calibration transitions.

\textbf{Theorem 9 (confusability floor).} For any $b$ with $\|b\|^{d+2} \lesssim \sigma^2 (L\Delta)^{d} / (cN)$ there exist $u \in \mathcal{U}(L)$, a state $z$, and $u' \in \mathcal{U}(L)$ agreeing with $u$ outside a neighbourhood of $z$, such that the joint law of calibration and audited data under (a) drift $u$ with a single corruption of displacement $b$ at the operation visiting $z$, so $I_\delta = \{t\}$, and (b) drift $u'$ with no corruption, so $I_\delta = \varnothing$, differ by at most $1/4$ in total variation. Every auditor then satisfies $\mathbb{E}_a|\psi \triangle I_a| + \mathbb{E}_b|\psi \triangle I_b| \ge 3/4$.

Combining with the noise-limited boundary for a corruption at an unknown one of $T$ locations, the smallest attributable corruption satisfies
\[\rho^2_{\min} \;\asymp\; \sigma^2\sqrt{2 d \log T} \;+\; \rho^2_{\mathrm{floor}}, \qquad \rho_{\mathrm{floor}} \asymp \left(\frac{\sigma^2 (L\Delta)^d}{c\,nT}\right)^{\frac{1}{d+2}}.\]

\textbf{Corollary 7.} The floor decays in the number of valid trajectories as $n^{-1/(d+2)}$; halving it requires $2^{d+2}$ times more calibration data.

\begin{center}\begin{tabular}{rrrr}\hline
$d$ & floor after $100\times$ data & after $10^4\times$ & data multiple to halve the floor \\ \hline
2 & 0.32 & 0.10 & 16 \\
8 & 0.63 & 0.40 & $1.0\times10^{3}$ \\
32 & 0.87 & 0.76 & $1.7\times10^{10}$ \\
128 & 0.965 & 0.932 & $10^{39}$ \\
256 & 0.982 & 0.965 & $10^{78}$ \\
\hline\end{tabular}\end{center}

At $d=256$, a 100-fold increase in valid trajectories reduces this floor by less than 2\%.

\textbf{Corollary 8.} The dependence on $d$ is through the exponent, so reducing the dimension in which the dynamics are modeled changes the rate rather than the constant. From $d = 256$ to $d = 8$ turns a floor effectively fixed in $n$ into one decaying as $n^{-1/5}$.

This rate provides a stronger motivation for dimension reduction than the finite-dimensional score comparison in Section 4.2. The relevant dimension is the intrinsic dimension of the fitted dynamics, not necessarily the width of the embedding. A 256-dimensional embedding concentrated near an 8-dimensional manifold can have the more favorable rate if the model is fitted in a representation that exposes that structure. The intrinsic dimension of agent trajectories must be evaluated empirically.

\textbf{Corollary 9.} As $d$ grows with the other quantities fixed, $\rho_{\mathrm{floor}} \to L\Delta$: no auditor can distinguish a corruption from a legitimate feature of the dynamics when the corruption is smaller than what the dynamics may do over one operation.

The floor is an identifiability limit rather than an estimation-error term. For an auditor trained only on valid trajectories, a sufficiently small local corruption cannot be distinguished from an unobserved feature of the valid dynamics. Theorem 9 therefore limits the magnitude of departures that can be attributed, while the error-control results in Section 5 remain valid for the resulting scores.

\subsection{Adversarially chosen corruption}

The corruptions in Section 3.1 are exogenous. If the location of a perturbation is selected to maximize terminal effect while limiting detectability, its position along the trajectory becomes relevant.

Measure damage by terminal displacement. In the noiseless limit, a shift $b$ at normalized time $s_0$ produces terminal displacement $\|b\|(1-\bar{s})/(1-s_0)$ and detection noncentrality $\lambda=\|b\|^2/\sigma_{\mathrm{eff}}^2(s_0)$. Define the damage-to-detectability ratio $\mathcal{E}(s_0)=\sigma_{\mathrm{eff}}(s_0)(1-\bar{s})/(1-s_0)$.

\textbf{Proposition 15 (late shifts have greater damage-to-detectability ratio).} With $\sigma_{\mathrm{eff}}$ constant, $\mathcal{E}$ is strictly increasing in $s_0$. At $\bar{s}=0.9$, a shift at $s_0=0.75$ is four times as efficient as the same shift at the start of the trajectory, and a shift at $s_0=0.89$ is nine times as efficient.

Early shifts are contracted toward the anchor by the factor $(1-\bar{s})/(1-s_0)$, whereas late shifts are less attenuated. With constant $\sigma_{\mathrm{eff}}$, innovation detectability does not depend on position. The bridge therefore reduces terminal effects of early perturbations more strongly than those of late perturbations.

\textbf{Corollary 10.} Proposition 10 places reduced terminal sensitivity and Proposition 8 places high anchor leverage in the final operations. Under the stated model, the single shift with the largest terminal effect relative to detectability therefore occurs near the end of the trajectory.

A second strategy distributes the perturbation across operations. A displacement $b$ spread evenly over $k$ consecutive operations has noncentrality $\lambda^{(h)}(k)=\|b\|^2\min(h,k)^2/(k^2h\sigma^2)$ in the $h$-step channel, which is maximized at $h=k$. Relative to this matched horizon, a one-step score loses a factor $k$ in noncentrality.

\textbf{Theorem 10 (dyadic horizon ladder).} With horizons $\mathcal{H}=\{1,2,4,\ldots,2^J\}$ and a corruption spread over any $k\le2^J$ operations,
\[\max_{h \in \mathcal{H}} \lambda^{(h)}(k) \;\ge\; \tfrac{1}{\sqrt{2}}\,\lambda^{(k)}(k),\]
with equality at $k=\sqrt{2}\,2^j$. The number of tested horizons is $J+1=O(\log T)$.

The dyadic horizon set is therefore minimax within a factor $\sqrt{2}$ of the matched horizon, compared with a factor-$k$ loss for one-step scoring. The most difficult perturbations under the bridge model are gradual displacements concentrated in the final operations. Examples include progressive narrowing of a cohort definition, gradual relaxation of inclusion criteria, or drift in the selected covariate set. These results further motivate a multimodal terminal predictive in Section 8.

\section{Limitations}

Three limitations are structural.

First, the flagged set has no causal interpretation. The set $I$ contains mechanism violations but does not distinguish an initiating error from an independent later error or identify a causal chain among flagged operations.

Second, assumption (A2a) may fail when the agent's mechanism changes after an error. Sustained procedural changes invalidate the benign-step model even if individual downstream operations are locally coherent. The total-variation bound in Theorem 5 can then approach its trivial limit, and off-support variance inflation in Corollary 4 does not address a changed transition mechanism. A latent-regime model is more appropriate in this setting: the estimand becomes the time at which the process leaves the valid regime, yielding a changepoint problem with an unknown post-change law. E-detectors provide non-asymptotic guarantees for this setting (Shin, Ramdas and Rinaldo, 2024), and conformal test martingales provide an exchangeability-based alternative (Vovk, Nouretdinov and Gammerman, 2003). Multi-horizon residuals address bounded gradual perturbations, but not a persistent regime change.

Third, report-specific false-discovery guarantees are weak at the trajectory lengths considered here. Proposition 6 bounds the realized false discovery proportion simultaneously over thresholds, but the additive term is large when $T$ is in the tens. A reader therefore receives a long-run FDR guarantee and a realized point estimate, not a high-probability statement for the individual report.

The identifiability results in Section 11 impose two additional limits. The exact corrupted set is not testable without a minimum-effect threshold, and the smallest attributable departure has a confusability floor that decreases extremely slowly with calibration sample size in high dimension.

Accordingly, the strongest guarantees apply when a departure occurs near the training support and the post-error transition mechanism remains within the modeled family.

\appendix

\section{Proofs}

Throughout, $\{T \ge t\} \in \mathcal{F}_{t-1}$ because the decision to continue after operation $t-1$ is made from the realized history. Part I invokes (A2a) at on-support benign steps. Off-support benign steps are treated through (A2b) and Theorem 7, which provides tail dominance rather than equality and preserves results that require only super-uniformity.

\subsection{Theorem 1}

Write $E_t = \{t \le T,\ t \notin I\}$. Since $\{T \ge t\} \in \mathcal{F}_{t-1} \subseteq \mathcal{G}_t$ and $\{t \in I\} \in \mathcal{G}_t$ by (A3), we have $E_t \in \mathcal{G}_t$. The parameters of $Q_t$ are $\mathcal{G}_t$-measurable, so $R_t$ is a $\mathcal{G}_t$-measurable map; under (A2a) and (A0), $\mathcal{L}(x_t \mid \mathcal{G}_t) = Q_t$ is atomless on $E_t$, and the Rosenblatt transform of a law by itself is uniform (Rosenblatt, 1952), giving $\mathcal{L}(u_t \mid \mathcal{G}_t) = \mathrm{Unif}[0,1]^d$ on $E_t$. Hence

\[\mathbb{E}[f(u_t)\mathbf{1}_{E_t}] = \mathbb{E}\big[\mathbf{1}_{E_t}\,\mathbb{E}[f(u_t) \mid \mathcal{G}_t]\big] = \mathbb{E}[f(U)]\,\mathbb{P}(E_t).\]

For $t_1 < \cdots < t_r$, the variables $u_{t_1}, \ldots, u_{t_{r-1}}$ and the event $\bigcap_{j<r}E_{t_j}$ are measurable with respect to $\mathcal{F}_{t_r-1}\subseteq\mathcal{G}_{t_r}$. Conditioning on $\mathcal{G}_{t_r}$ separates the last factor, and iteration gives the product identity. The argument does not condition on the realized value of $T$ or $I$, which permits a stopping time and random predictable corruption set.

For (ii), on $\{t \in I\}$ the conditional law of $x_t$ given $\mathcal{G}_t$ is $\tilde{Q}_t$ by definition, so $u_t \mid \mathcal{G}_t \sim R_{t\#}\tilde{Q}_t$. Corollary 1 follows because $\varepsilon_t = \nu_t + \xi_t$ with $\nu_t$ being $\mathcal{G}_t$-measurable, and the monotonicity of the noncentral chi-square tail in its noncentrality is classical (Johnson, Kotz and Balakrishnan, 1995, Ch. 29). Corollary 2 follows by diagonalizing $\Gamma_t$; if $\Gamma_t \prec 0$ then $\sum_i(1+\gamma_{t,i})\chi^2_1$ is stochastically smaller than $\chi^2_d$, so the upper-tail rejection probability is below $\alpha$. $\square$

\subsection{Theorem 2}

Let $D(s) = \tilde{z}(s) - z(s)$, so $D' = -D/(1-s) + [u_\theta(\tilde{z},s) - u_\theta(z,s)]$. Wherever $D \ne 0$,
\[\frac{d}{ds}\|D\| = -\frac{\|D\|}{1-s} + \big\langle D/\|D\|,\ u_\theta(\tilde{z},s) - u_\theta(z,s)\big\rangle,\]
and the inner product is bounded in modulus by $L\|D\|$ under (A4). Hence
\[-\frac{1}{1-s} - L \;\le\; \frac{d}{ds}\log\|D(s)\| \;\le\; -\frac{1}{1-s} + L.\]
Integrating from $s_0$ and using $\int_{s_0}^s (1-v)^{-1}dv = \log\frac{1-s_0}{1-s}$ gives the envelopes. The lower envelope is strictly positive on $[s_0,\bar{s}]$, so $D$ never vanishes and the differentiation is justified; at isolated points of non-differentiability of $\|D\|$ the same bounds hold for the Dini derivatives. If $\Sigma$ is state-independent, running both stochastic dynamics with the same noise cancels the noise in the difference and the envelope holds pathwise. The lower envelope decays for every $L$, since $\frac{d}{ds}\log[(1-s)e^{-L(s-s_0)}] = -1/(1-s) - L < 0$; the upper envelope decays only when $L < 1/(1-s)$. For $L$ above that the two envelopes separate and the bound determines only that the gap is nonzero, not whether it decays. For repeated shifts $b_i$ at times $s_i$, superposition of the upper envelope gives $\|D(s)\| \le \sum_{s_i \le s}\|b_i\|\frac{1-s}{1-s_i}e^{L(s-s_i)}$; the lower bound does not superpose, since shifts can cancel. $\square$

\subsection{Propositions 1 and 2}

\emph{Proposition 1(i).} Fix $t \ge t_0 + k$. The conditioning $\sigma$-field of $Q^{(k)}_t$ contains $x_{t-k}$ with $t-k \ge t_0$, so the corrupted transition at $t_0$ lies in the past of the prediction. Every transition from $t-k+1$ to $t$ is benign and follows $\mathcal{M}$ under (A2a), so $Q^{(k)}_t$ is the correct conditional law and the residual is uniform by the argument of Appendix A.1 applied to the coarser filtration. For $t_0 \le t \le t_0 + k - 1$ we have $t - k < t_0$, so the corrupted transition lies strictly inside the prediction window and enters the predictive mean.

\emph{Proposition 1(ii).} The conditioning $\sigma$-field of $Q_t^{(k)}$ is $\mathcal{G}^{(k)}_t = \mathcal{F}_{t-k} \vee \sigma(c_{t-k+1}, \ldots, c_t)$. The argument of Appendix A.1 requires the benign event to lie in that field. Now $\{t \notin I\} \in \mathcal{G}_t$ need not lie in $\mathcal{G}^{(k)}_t$, and $\{t \le T\}$ depends on the halting decisions at $t-k+1, \ldots, t-1$ and hence on $x_{t-k+1}, \ldots, x_{t-1}$, which are not in $\mathcal{F}_{t-k}$. Under the stated commitment condition both events lie in $\mathcal{G}^{(k)}_t$ and Appendix A.1 applies unchanged. Without it, conditioning on survival to $t$ selects on the intervening innovations and the residual is not uniform.

\emph{Proposition 1(iii).} If $t_{j+1} - t_j \ge k$ then $u^{(k)}_{t_j}$ is $\mathcal{F}_{t_j}$-measurable and $\mathcal{F}_{t_j} \subseteq \mathcal{F}_{t_{j+1}-k} \subseteq \mathcal{G}^{(k)}_{t_{j+1}}$, so the peeling argument of Appendix A.1 goes through. If $t_{j+1} - t_j < k$ then $x_{t_j}$ lies strictly inside the prediction window of $t_{j+1}$ and the two residuals share $k - (t_{j+1}-t_j)$ innovations, so they are dependent; nothing above bounds that dependence. $\square$

\emph{Proposition 2.} With $\Sigma \equiv \sigma^2 I$ and identity drift over the window, $x_t - \mathbb{E}[x_t \mid \mathcal{F}_{t-k}] = \sum_{j=t-k+1}^{t}(b_j + \sigma\xi_j)$, which has mean $\sum_j b_j$ and covariance $k\sigma^2 I$. At $t = t_0 + k - 1$ the accumulated shift is $b$, so the whitened mean has squared norm $\|b\|^2/(k\sigma^2)$. The one-step residual at any single step of the window has whitened mean $b/(k\sigma)$, giving $\|b\|^2/(k^2\sigma^2)$. $\square$

\subsection{Propositions 3, 4 and 5}

\emph{Proposition 3.} Write $a = \sum_{i}(\nu_i + \xi_i)^2 = d + \lambda_d + S_1 + S_2$ with $S_1 = \sum_i(\xi_i^2 - 1)$ and $S_2 = 2\sum_i\nu_i\xi_i$. Then $S_1/\sqrt{2d} \Rightarrow \mathcal{N}(0,1)$, $S_2 \sim \mathcal{N}(0, 4\lambda_d)$ exactly, and $\mathrm{Cov}(S_1,S_2) = 0$ since $\mathbb{E}[\xi^3] = 0$. When $\lambda_d = O(\sqrt{d})$, $S_2/\sqrt{2d}$ has standard deviation $\sqrt{2\lambda_d/d} \to 0$, so $(a-d)/\sqrt{2d} = \lambda_d/\sqrt{2d} + S_1/\sqrt{2d} + o_P(1)$ while $(\chi^2_{d,1-\alpha} - d)/\sqrt{2d} \to z_\alpha$. If $\lambda_d/\sqrt{2d} \to \infty$, then $a$ has mean $d + \lambda_d$ and standard deviation $\sqrt{2d + 4\lambda_d} = o(\lambda_d) + O(\sqrt{d})$, and Chebyshev gives power tending to one. $\square$

\emph{Proposition 4.} The Bonferroni test has level at most $\alpha$ by the union bound. For power, fix a signal coordinate: $\mathbb{P}(|\delta_d + \xi| > z_{\alpha/(2d)}) \ge \mathbb{P}(\xi > z_{\alpha/(2d)} - \delta_d)$, and since $z_{\alpha/(2d)} \le \sqrt{2\log(2d/\alpha)} = \sqrt{2\log d}(1+o(1))$ while $\delta_d = \sqrt{2(1+\eta)\log d}$, the gap diverges. For the chi-square test, $\lambda_d = k_d\delta_d^2 = 2(1+\eta)k_d\log d = o(\sqrt{d})$ by hypothesis, and Proposition 3 applies. The finite-dimensional entries in the table of Section 4.2 are exact: $\lambda^\ast$ solves $\mathbb{P}(\chi^2_d(\lambda) > \chi^2_{d,0.99}) = 0.8$, and $\lambda^\ast_k = k\delta^2$ where $\delta$ solves $1 - [\Phi(c-\delta)-\Phi(-c-\delta)]^k[2\Phi(c)-1]^{d-k} = 0.8$ with $c$ the exact level-$\alpha$ maximum threshold, $2(1-\Phi(c)) = 1 - (1-\alpha)^{1/d}$. $\square$

\emph{Proposition 5.} If $\Sigma$ is diagonal so is $\Sigma^{-1/2}$, and supports are preserved. Otherwise take $\Sigma^{-1/2}$ with a dense first column and $b = e_1$; then $\Sigma^{-1/2}b$ is that column. $\square$

\subsection{Lemma 1, Theorems 3 and 4}

Under (A1$^\ast$) the calibration scores are i.i.d.\ from the null score law; under (A2a) and Theorem 1 each benign test score has the same law and is independent of the calibration set, which comes from distinct trajectories. The $n+1$ scores are exchangeable with continuous common law, so the rank of $a_t$ is uniform and $\mathbb{P}(p_t \le u) \le u$ (Vovk, Gammerman and Shafer, 2005). Under (A1) alone the same argument runs within a stratum, where exchangeability of the test score with the calibration scores at that index is assumed directly rather than derived from correct specification. (A5) is what permits the statement to be read conditionally on the trajectory having been submitted for audit.

For Theorem 3, pad with $p_t = 1$ for $T < t \le T_{\max}$. Padded p-values satisfy $\mathbb{P}(p_t \le u) = 0$ for $u < 1$, so the padded family has $T_{\max}$ hypotheses of which the padded ones are super-uniform, and the count is now deterministic. Benjamini--Yekutieli requires only marginal super-uniformity of the nulls and controls FDR at $q\,m_0/T_{\max}$ under arbitrary dependence (Benjamini and Yekutieli, 2001, Theorem 1.3). A padded index is rejected only if $1 \le q k /(T_{\max}H_{T_{\max}})$, which fails for all $k \le T_{\max}$ when $q < H_{T_{\max}}$, so the realized discoveries are unchanged.

For Theorem 4, the likelihood ratio $f_{d,\lambda}(a)/f_d(a)$ has expectation one under $f_d$ for each $\lambda$, hence so does any mixture; the closed form follows from the Poisson-mixture representation of the noncentral density, $f_{d,\lambda}(a) = e^{-\lambda/2}\sum_{j\ge0}\frac{(\lambda/2)^j}{j!}f_{d+2j}(a)$ together with $f_{d+2j}(a)/f_d(a) = \frac{\Gamma(d/2)}{\Gamma(d/2+j)}(a/2)^j$, so that the ratio is $e^{-\lambda/2}\,{}_0F_1(;d/2;\lambda a/4)$ and the Bessel form is the standard identity for ${}_0F_1$. For a calibrated e-value, if $f$ is decreasing with $\int_0^1 f = 1$ and $p$ is super-uniform, then $\mathbb{E}[f(p)] \le \mathbb{E}[f(U)] = 1$. e-BH controls FDR at $q\,m_0/T_{\max}$ under arbitrary dependence (Wang and Ramdas, 2022). $\square$

\subsection{Proposition 6}

Under (A1$^\ast$) and (A2a) the exact-null p-values $p_t = \mathbb{P}(\chi^2_d \ge a_t)$ are i.i.d.\ uniform across benign steps by Theorem 1. Let $V(s)$ count benign p-values at most $s$. By the Dvoretzky--Kiefer--Wolfowitz inequality with Massart's constant, with probability at least $1-\delta$, $\sup_s |V(s)/m_0 - s| \le \sqrt{\log(2/\delta)/(2m_0)}$, so $V(s) \le m_0 s + \sqrt{m_0\log(2/\delta)/2}$ simultaneously in $s$; bounding $m_0 \le T$ and dividing by $R(s)\vee1$ gives the claim.

With conformal rather than exact-null p-values the benign p-values are dependent through the shared calibration set. Conditioning on it restores independence, and a second application of the same inequality to the calibration sample gives $\mathbb{P}(p_t \le s \mid \Omega) \le s + \sqrt{\log(2/\delta')/(2n)}$ uniformly in $s$ with probability $1-\delta'$; the bound then holds with $s$ replaced by $s + \sqrt{\log(2/\delta')/(2n)}$ and total failure probability $\delta + \delta'$. $\square$

\subsection{Theorem 5 and Proposition 7}

\emph{Theorem 5.} Construct both worlds on one space, sequentially in $t$. Given agreement of histories through $t-1$: if $t \in I$, both worlds apply the same corruption kernel to the same history and can be coupled to agree exactly; if $t \notin I$, maximally couple the true conditional law with $Q_t$, which fail to agree with probability at most $\delta_t$ (Lindvall, 2002). If the histories already disagree, couple arbitrarily. Let $E$ be the event of agreement through $T$; a union bound over benign steps gives $\mathbb{P}(E^c) \le \sum_{t\notin I}\delta_t$. The calibration set is shared and exact. On $E$ the two worlds produce identical scores, flags and corrupted sets, hence identical false discovery proportions. Since $\mathrm{FDP} \le 1$,
\[\mathrm{FDR}_{\mathrm{true}} = \mathbb{E}[\mathrm{FDP}\,\mathbf{1}_E] + \mathbb{E}[\mathrm{FDP}\,\mathbf{1}_{E^c}] \le \mathrm{FDR}_{\mathrm{model}} + \mathbb{P}(E^c). \qquad \square\]

\emph{Proposition 7.} Let $\hat{S}_n(a) = n^{-1}\#\{\omega_j \ge a\}$, let $S_G$ be the survival function of the true calibration score law and $S$ that of the model null. Then $p_t = \frac{n}{n+1}\hat{S}_n(a_t) + \frac{1}{n+1}$, so $|p_t - S(a_t)| \le \|\hat{S}_n - S\|_\infty + \frac{1}{n+1}$, and $\|\hat{S}_n - S\|_\infty \le \|\hat{S}_n - S_G\|_\infty + \epsilon$. The Dvoretzky--Kiefer--Wolfowitz inequality bounds the first term by $\sqrt{\log(2/\delta)/(2n)}$ with probability $1-\delta$. On that event $\tilde{p}_t = p_t + \eta \ge S(a_t)$, which is uniform under (A2a) at the test step, so $\tilde{p}_t$ is super-uniform and every procedure valid for super-uniform nulls applies, with $\delta$ added to the error rate to account for the complement. $\square$

\subsection{Theorem 7, Corollary 4 and Proposition 9}

Write $x - \mu_m = (\mu^\ast - \mu_m) + C^{1/2}\zeta$ with $\|\mu^\ast - \mu_m\| \le \kappa D + \eta_0$, $C \preceq \bar\sigma^2 I$ and $\zeta \sim \mathcal{N}(0,I_d)$. Pointwise, $\|x-\mu_m\|^2 \le 2(\kappa D + \eta_0)^2 + 2\bar\sigma^2\|\zeta\|^2$, so with $A(D) = 2(\kappa D+\eta_0)^2/\sigma^2(D)$ and $B(D) = 2\bar\sigma^2/\sigma^2(D)$,
\[a_t \le A(D) + B(D)\,\chi^2_d \quad \text{pointwise in } \zeta.\]
For $D \ge d_0$, $A(D) \le 2(\kappa D + \eta_0)^2/(\sigma_0^2\gamma D^2) \le 2(\kappa + \eta_0/d_0)^2/(\sigma_0^2\gamma) = A_\infty$ and $B(D) \le B(d_0) = B_0$. Hence for any $u$,
\[\mathbb{P}(a_t > u) \le \mathbb{P}\big(A_\infty + B_0\chi^2_d > u\big) = \mathbb{P}\big(\chi^2_d > (u - A_\infty)/B_0\big),\]
which is at most $\mathbb{P}(\chi^2_d > u)$ precisely when $(u-A_\infty)/B_0 \ge u$, that is $u \ge A_\infty/(1-B_0)$. Both $A_\infty$ and $B_0$ are decreasing in $\gamma$, so the two conditions $B_0 < 1$ and $A_\infty/(1-B_0) \le \chi^2_{d,1-\alpha}$ hold for all $\gamma$ above an explicit threshold. Super-uniformity at levels $u \le \alpha$ follows because $p_t \le u \le \alpha$ corresponds to $a_t$ exceeding a threshold at least $\chi^2_{d,1-\alpha}$. If the true deviation is $\bar\sigma^2$-sub-Gaussian rather than Gaussian, replace $\|\zeta\|^2$ by its sub-Gaussian envelope $(\sqrt{d}+t)^2$ with failure probability $e^{-t^2/2}$; the conclusion holds with $\chi^2_d$ replaced by that envelope and an additive $e^{-t^2/2}$.

Corollary 4 is the special case in which the state is off-support because it inherits a corruption. Proposition 9 is immediate: $\varepsilon = b/\sigma(D) + \xi$ gives $a \sim \chi^2_d(\|b\|^2/\sigma^2(D))$, and Corollary 3 converts the noncentrality into a requirement on $\|b\|$. $\square$

\subsection{Proposition 8}

Take $\varrho = 0$ for legibility; the general case replaces $W$ by $W+\varrho$ throughout. With $\varepsilon_t = \xi_t + v_t g$, $\hat{g} = W^{-1}\sum_s v_s\varepsilon_s = g + W^{-1}\sum_s v_s\xi_s$, so $\tilde{\varepsilon}_t = \xi_t - v_t W^{-1}\sum_s v_s\xi_s$ is free of $g$. Its covariance is
\[\big(1 - 2v_t^2/W + v_t^2W/W^2\big)I_d = (1-h_t)I_d, \qquad h_t = v_t^2/W,\]
and $\mathrm{Cov}(\tilde{\varepsilon}_t,\tilde{\varepsilon}_s) = -v_t v_s W^{-1}I_d$ for $s \ne t$, which is the cross-correlation in part (i). Hence $\tilde{a}_t = \|\tilde{\varepsilon}_t\|^2/(1-h_t) \sim \chi^2_d$ marginally when no step is corrupted. If a shift $\nu_t$ is present at step $t$ only, $\mathbb{E}[\tilde{\varepsilon}_t] = (1-h_t)\nu_t$ and $\mathbb{E}[\tilde{\varepsilon}_s] = -v_sv_t W^{-1}\nu_t$, so the standardized residual at $s$ is $\mathcal{N}(\theta_s, I_d)$ with $\|\theta_s\|^2 = \mu_s = h_s h_t\lambda_t/(1-h_s)$.

For the resulting bound, $\mathrm{TV}(\mathcal{N}(\theta,I_d),\mathcal{N}(0,I_d)) = 2\Phi(\|\theta\|/2)-1 \le \|\theta\|/\sqrt{2\pi}$, and total variation between the score laws is no greater than total variation between the residual laws. Summing $\sqrt{\mu_s}=\sqrt{h_t\lambda_t}\sqrt{h_s/(1-h_s)}$ over $s\ne t$ gives the displayed bound, which grows as $\sqrt{\lambda_t}$. Under a bounded-influence estimator with influence bounded by $\varsigma$, the induced mean satisfies $\|\theta_s\|\le v_s\varsigma/\sqrt{1-h_s}$ independently of $\lambda_t$, and the sum is uniformly bounded in the corruption magnitude. $\square$

\subsection{Proposition 10}

With the predictable anchor $g_\eta$ of Lipschitz constant $L_g$ and $h_T = 1-s_{T-1} \ge \Delta_T$, the terminal model mean satisfies
\[\mu(z) - \mu(z') = (z-z')\Big(1 - \frac{\Delta_T}{h_T}\Big) + \frac{\Delta_T}{h_T}\big(g_\eta(z) - g_\eta(z')\big) + \Delta_T\big(u_\theta(z) - u_\theta(z')\big),\]
so $\|\mu(z)-\mu(z')\| \le w\,\kappa_T$ with $\kappa_T = 1 + L_g\Delta_T/h_T + L\Delta_T$. Since $\|m_1-m_2\| \ge 2r$, the triangle inequality implies $\max_i\|\mu(z_i)-m_i\| \ge r - w\kappa_T/2 = \rho$. At the state with the larger discrepancy, $\|x-\mu\| \ge \rho - \sigma_c\|\zeta\|$, so the flag probability is at least $\mathbb{P}(\sigma_c\|\zeta\| < \rho - \sigma_T\sqrt{\tau})$. If alternative (b) fails, then $\sigma_T\sqrt{\tau}<\rho/2$, and $\rho \ge 2\sigma_c(\sqrt{d}+t)$ gives a probability of at least $1-e^{-t^2/2}$ by Gaussian norm concentration. If instead $\sigma_T^2 \ge \rho^2/(4\tau)$, a terminal shift has $\lambda_{\mathrm{eff}}=\|b\|^2/\sigma_T^2 \le 4\tau\|b\|^2/\rho^2$. For $\tau\le2d$ at large $d$, the noncentrality reaches the $\sqrt{2d}$ boundary of Proposition 3 only when $\|b\|^2 \ge \rho^2\sqrt{2d}/(8d)$, which gives $c_1$.

The hypothesis $\rho > 0$ is what makes the dichotomy bind, and it fails when
\[L_g \;>\; \Big(\frac{2r}{w} - 1 - L\Delta_T\Big)\frac{h_T}{\Delta_T}.\]
An anchor head whose state sensitivity exceeds this threshold can map the two states to different terminal means, so the hypothesis of Proposition 10 no longer holds. This capacity condition does not establish that either alternative is avoided; accurate routing is addressed in Proposition 12. The threshold increases with $h_T/\Delta_T$, making separation most difficult when the final operation has low cost while substantial budget remains. $\square$

\subsection{Theorem 6 and Proposition 13}

\emph{Theorem 6.} The odds bound gives, for any valid $\tau, \tau'$, $\pi(\tau)\big(1-\pi(\tau')\big) \le \Gamma\,\pi(\tau')\big(1-\pi(\tau)\big)$, hence $\pi(\tau) \le \Gamma\pi(\tau') + \pi(\tau)\pi(\tau')(1-\Gamma) \le \Gamma\pi(\tau')$ since $\Gamma \ge 1$. Condition on the calibration set $\Omega$, which is drawn without selection. For any event $A$ measurable in the audited trajectory,
\[\mathbb{P}(A \mid S, \Omega) = \frac{\mathbb{E}[\pi(\tau)\mathbf{1}_A \mid \Omega]}{\mathbb{E}[\pi(\tau) \mid \Omega]} \le \frac{\sup \pi}{\inf \pi}\,\mathbb{P}(A \mid \Omega) \le \Gamma\,\mathbb{P}(A \mid \Omega).\]
Taking $A = \{p_t \le u\}$ for a benign $t$ and averaging over $\Omega$ gives $\mathbb{P}(p_t \le u \mid S) \le \Gamma u$ by Lemma 1. Substituting the inflated super-uniformity into the Benjamini--Yekutieli or e-BH argument multiplies the bound by $\Gamma$, so running at $q/\Gamma$ restores level $q$. The expression for $\Gamma^\ast$ is the largest $\Gamma$ for which the realized $p_{(k)}$ still satisfies the Benjamini--Yekutieli threshold at level $q/\Gamma$. $\square$

\emph{Proposition 13.} Under (A2a) and correct specification, the occurrences of type $c$ in distinct trajectories are independent, each contributing $\chi^2_d(\lambda)$ by Corollary 1, so the sum is $\chi^2_{m_c d}(m_c\lambda)$ by additivity of independent noncentral chi-squares. Proposition 3 applied in dimension $m_c d$ requires noncentrality of order $\sqrt{2m_c d}\,(z_\alpha + z_{1-\beta})$; setting $m_c\lambda$ equal to that and dividing by $m_c$ gives the stated bound. For Corollary 5, the confusability construction in Appendix A.14 perturbs the drift of the \emph{valid} mechanism and is distinguished only by calibration transitions passing through the perturbed region; audited occurrences are draws from the corrupted mechanism and enter neither side of that comparison. $\square$

\subsection{Propositions 11 and 12}

\emph{Proposition 11.} The terminal model mean is $\mu(z) = z(1 - \Delta_T/h_T) + (\Delta_T/h_T)g_\eta(z) + \Delta_T u_\theta(z)$, so $\mathrm{Lip}(\mu) \le 1 + L_g\Delta_T/h_T + L\Delta_T = \kappa_T$, and the extrapolation bound of Theorem 7 holds with $\kappa \le \kappa_T + L^\ast$. Solving the second design condition of Theorem 7, $A_\infty/(1-B_0) \le \chi^2_{d,1-\alpha}$ with $A_\infty = 2(\kappa + \eta_0/d_0)^2/(\sigma_0^2\gamma)$, for $\gamma$ gives the first display. Substituting into $\|b\|^2_{\min} = \sigma_0^2(1+\gamma D^2)\sqrt{2d}(z_\alpha + z_{1-\beta})$ of Proposition 9 and dropping the additive one gives the second, with $\sigma_0^2$ cancelling. If $L_g > (2r/w - 1 - L\Delta_T)h_T/\Delta_T$ then $L_g\Delta_T/h_T > 2r/w - 1 - L\Delta_T$, so $\kappa_T > 2r/w$. $\square$

\emph{Proposition 12.} At the terminal step, $x - \mu(z) = (m(z) - \mu(z)) + \sigma_c\zeta$ with $\|m(z)-\mu(z)\| \le e_0$ and $\zeta \sim \mathcal{N}(0,I_d)$, so pointwise $a \le A + B\chi^2_d$ with $A = 2e_0^2/\sigma_T^2$ and $B = 2\sigma_c^2/\sigma_T^2$. This has the form used in Theorem 7, and the argument of Appendix A.8 gives $\mathbb{P}(a > u) \le \mathbb{P}(\chi^2_d > u)$ for $u \ge A/(1-B)$. The conditions $B < 1$ and $A/(1-B) \le \chi^2_{d,1-\alpha}$ are jointly equivalent to the stated lower bound on $\sigma_T^2$. Detectability follows from Proposition 9 with $\sigma^2(D)$ replaced by $\sigma_T^2$. Alternative (b) of Proposition 10 gives blindness up to $\|b\|^2 \le c_1\rho^2/\sqrt{d}$, whereas accurate routing gives detectability from $\|b\|^2 \gtrsim \sigma_c^2\sqrt{2d}$. Their ratio is of order $\rho^2/(\sigma_c^2d)$ and is at least four when $\rho \ge 2\sigma_c\sqrt{d}$. $\square$

\subsection{Theorem 8 and Corollary 6}

Let $K^{(1)}$ agree with $K^\ast$ except at $t_0$, where it equals an arbitrary $Q \ne K^\ast_{t_0}$. Let $K^{(2)}$ agree with $K^{(1)}$ at $t_0$ and, for $t > t_0$, take $K^{(2)}_t = (1-\epsilon)K^\ast_t + \epsilon R_t$ with $R_t \ne K^\ast_t$ arbitrary. Every $t > t_0$ satisfies $K^{(2)}_t \ne K^\ast_t$, so $I(K^{(2)}) = \{t_0,\ldots,T\}$, while $\mathrm{TV}(K^{(2)}_t, K^\ast_t) \le \epsilon$ uniformly; sequential maximal coupling gives $\mathrm{TV}(P^{(1)},P^{(2)}) \le (T-t_0)\epsilon$, and $\epsilon = \eta/(T-t_0)$ gives the bound. Symmetric difference obeys the triangle inequality, so for every set $S$, $|S \triangle I^{(1)}| + |S \triangle I^{(2)}| \ge |I^{(1)} \triangle I^{(2)}| = T-t_0$ pointwise, whence
\[\mathbb{E}_1|\psi \triangle I^{(1)}| + \mathbb{E}_2|\psi \triangle I^{(2)}| \ge (T-t_0)\int \min(dP^{(1)},dP^{(2)}) \ge (T-t_0)(1-\eta).\]
Corollary 6 follows since $\eta$ is arbitrary and neither the sample size nor the score enters the bound. $\square$

\subsection{Theorem 9}

Fix $z$ in the interior of the visited region and let $\phi$ be a smooth bump supported on the ball of radius $h$ about $z$ with $\|\phi\|_\infty = 1$ and Lipschitz constant of order $1/h$. Put $\beta = (b/\Delta)\phi$ and $u' = u + \beta$; membership in $\mathcal{U}(L)$ requires $h \gtrsim \|b\|/(L\Delta)$, and $h$ is taken at that bound, which minimizes the calibration signal.

Under (a) the calibration trajectories have drift $u$ and under (b) drift $u'$. The laws differ only at transitions originating in $\mathrm{supp}\,\phi$, each contributing Kullback--Leibler divergence at most $\|b\|^2/(2\sigma^2)$. With $p = \mathbb{P}(X \in \mathrm{supp}\,\phi) \lesssim c h^d$, the chain rule and Pinsker's inequality give
\[\mathrm{TV}_{\mathrm{cal}} \lesssim \sqrt{\frac{c N \|b\|^{2}}{\sigma^2}\left(\frac{\|b\|}{L\Delta}\right)^{d}}.\]
Under the stated condition, this quantity is at most a small constant. Conditional on the audited trajectory visiting $\mathrm{supp}\,\phi$ once, at the step carrying the corruption in (a), the two audited laws coincide: in (a) that transition has mean $\mu_t + b$ by construction and in (b) mean $\mu_t + \beta(z)\Delta = \mu_t + b$ because the drift carries the bump, all other transitions following $u$ in both. An additional visit has probability at most $Tp$, negligible under the same condition when $T \lesssim n$. Summing and choosing constants gives $\mathrm{TV} \le 1/4$; since $|I_a \triangle I_b| = 1$, the argument in Appendix A.13 gives the risk bound. For the noise term, with $T$ candidate locations and known dynamics the maximum of $T$ independent $\chi^2_d$ scores exceeds $d + \sqrt{2d\log T}(1+o(1))$, so noncentrality of order $\sqrt{2d\log T}$ is required. $\square$

\subsection{Proposition 15 and Theorem 10}

\emph{Proposition 15.} In the noiseless bridge flow the gap between perturbed and unperturbed solutions is $\|D(s)\| = \|b\|(1-s)/(1-s_0)$ when the drift correction is absent, which is Theorem 2 with $L = 0$. Evaluating at $\bar s$ and dividing by $\sqrt{\lambda} = \|b\|/\sigma_{\mathrm{eff}}(s_0)$ gives $\mathcal{E}$. Monotonicity holds whenever $\sigma_{\mathrm{eff}}(s)(1-s)^{-1}$ is nondecreasing, in particular for constant $\sigma_{\mathrm{eff}}$. $\square$

\emph{Theorem 10.} Write $k=2^j t$ with $t\in[1,2)$. For ladder points $h\le k$, $\lambda^{(h)}(k)=\|b\|^2 h/(k^2\sigma^2)$ increases with $h$, so the maximizing choice is $h=2^j$. For $h\ge k$, $\lambda^{(h)}(k)=\|b\|^2/(h\sigma^2)$ decreases with $h$, so the maximizing choice is $h=2^{j+1}$. Relative to the oracle $\|b\|^2/(k\sigma^2)$, the ratio is $\max(2^j/k,k/2^{j+1})=\max(1/t,t/2)$, whose minimum over $t\in[1,2)$ is $1/\sqrt{2}$ at $t=\sqrt{2}$. Direct evaluation for $k\le2048$ gives the same worst-case ratio, attained near $k=181\approx\sqrt{2}\,128$. $\square$

\section*{References}

Basseville, M. and Nikiforov, I. V. (1993). \emph{Detection of Abrupt Changes: Theory and Application}. Prentice Hall, Englewood Cliffs, NJ.

Bates, S., Candès, E., Lei, L., Romano, Y., and Sesia, M. (2023). Testing for outliers with conformal p-values. \emph{The Annals of Statistics}, 51(1):149--178.

Benjamini, Y. and Hochberg, Y. (1995). Controlling the false discovery rate: a practical and powerful approach to multiple testing. \emph{Journal of the Royal Statistical Society, Series B}, 57(1):289--300.

Benjamini, Y. and Yekutieli, D. (2001). The control of the false discovery rate in multiple testing under dependency. \emph{The Annals of Statistics}, 29(4):1165--1188.

Cox, D. R. and Snell, E. J. (1968). A general definition of residuals. \emph{Journal of the Royal Statistical Society, Series B}, 30(2):248--265.

Dawid, A. P. (1984). Present position and potential developments: some personal views. Statistical theory: the prequential approach. \emph{Journal of the Royal Statistical Society, Series A}, 147(2):278--290.

Donoho, D. and Jin, J. (2004). Higher criticism for detecting sparse heterogeneous mixtures. \emph{The Annals of Statistics}, 32(3):962--994.

Doob, J. L. (1957). Conditional Brownian motion and the boundary limits of harmonic functions. \emph{Bulletin de la Société Mathématique de France}, 85:431--458.

Dvoretzky, A., Kiefer, J., and Wolfowitz, J. (1956). Asymptotic minimax character of the sample distribution function and of the classical multinomial estimator. \emph{The Annals of Mathematical Statistics}, 27(3):642--669.

Fan, J. (1996). Test of significance based on wavelet thresholding and Neyman's truncation. \emph{Journal of the American Statistical Association}, 91(434):674--688.

Feng, N., Sui, Y., Hou, S., Wu, G., and Cresswell, J. C. (2026). Conformal agent error attribution. arXiv:2605.06788.

Gertler, J. (1998). \emph{Fault Detection and Diagnosis in Engineering Systems}. Marcel Dekker, New York.

Grönwall, T. H. (1919). Note on the derivatives with respect to a parameter of the solutions of a system of differential equations. \emph{Annals of Mathematics}, 20(4):292--296.

Ingster, Y. I. and Suslina, I. A. (2003). \emph{Nonparametric Goodness-of-Fit Testing Under Gaussian Models}. Lecture Notes in Statistics 169. Springer, New York.

Johnson, N. L., Kotz, S., and Balakrishnan, N. (1995). \emph{Continuous Univariate Distributions, Volume 2}, 2nd ed. Wiley, New York.

Kailath, T. (1968). An innovations approach to least-squares estimation, Part I: linear filtering in additive white noise. \emph{IEEE Transactions on Automatic Control}, 13(6):646--655.

Katsevich, E. and Ramdas, A. (2020). Simultaneous high-probability bounds on the false discovery proportion in structured, regression, and online settings. \emph{The Annals of Statistics}, 48(6):3465--3487.

Kidger, P., Morrill, J., Foster, J., and Lyons, T. (2020). Neural controlled differential equations for irregular time series. \emph{Advances in Neural Information Processing Systems}, 33:6696--6707.

Kloeden, P. E. and Platen, E. (1992). \emph{Numerical Solution of Stochastic Differential Equations}. Springer, Berlin.

Lindvall, T. (2002). \emph{Lectures on the Coupling Method}. Dover, Mineola, NY.

Liu, J., Lin, Z., Padhy, S., Tran, D., Bedrax Weiss, T., and Lakshminarayanan, B. (2020). Simple and principled uncertainty estimation with deterministic deep learning via distance awareness. \emph{Advances in Neural Information Processing Systems}, 33:7498--7512.

Massart, P. (1990). The tight constant in the Dvoretzky--Kiefer--Wolfowitz inequality. \emph{The Annals of Probability}, 18(3):1269--1283.

Miyato, T., Kataoka, T., Koyama, M., and Yoshida, Y. (2018). Spectral normalization for generative adversarial networks. \emph{International Conference on Learning Representations}.

Page, E. S. (1954). Continuous inspection schemes. \emph{Biometrika}, 41(1--2):100--115.

Rogers, L. C. G. and Williams, D. (2000). \emph{Diffusions, Markov Processes and Martingales, Volume 2: Itô Calculus}, 2nd ed. Cambridge University Press, Cambridge.

Rosenblatt, M. (1952). Remarks on a multivariate transformation. \emph{The Annals of Mathematical Statistics}, 23(3):470--472.

Seitzer, M., Tavakoli, A., Antić, D., and Martius, G. (2022). On the pitfalls of heteroscedastic uncertainty estimation with probabilistic neural networks. \emph{International Conference on Learning Representations}.

Shin, J., Ramdas, A., and Rinaldo, A. (2024). E-detectors: a nonparametric framework for sequential change detection. \emph{New England Journal of Statistics in Data Science}, 2:229--260.

Tibshirani, R. J., Barber, R. F., Candès, E., and Ramdas, A. (2019). Conformal prediction under covariate shift. \emph{Advances in Neural Information Processing Systems}, 32.

Vovk, V., Gammerman, A., and Shafer, G. (2005). \emph{Algorithmic Learning in a Random World}. Springer, New York.

Vovk, V., Nouretdinov, I., and Gammerman, A. (2003). Testing exchangeability on-line. In \emph{Proceedings of the 20th International Conference on Machine Learning}, pp. 768--775. AAAI Press.

Vovk, V. and Wang, R. (2021). E-values: calibration, combination, and applications. \emph{The Annals of Statistics}, 49(3):1736--1754.

Wang, R. and Ramdas, A. (2022). False discovery rate control with e-values. \emph{Journal of the Royal Statistical Society, Series B}, 84(3):822--852.

Willsky, A. S. (1976). A survey of design methods for failure detection in dynamic systems. \emph{Automatica}, 12(6):601--611.

Yeh, S., Zhu, Y., Deep, S., and Li, S. (2026). Tracing agentic failure from the flow of success. arXiv:2607.12747.

Zhang, B., Zhu, J., Shi, Z., Liu, D., and Tang, R. (2026). AgentForesight: online auditing for early failure prediction in multi-agent systems. arXiv:2605.08715.

Zhang, S., Yin, M., Zhang, J., Liu, J., Han, Z., Zhang, J., Li, B., Wang, C., Wang, H., Chen, Y., and Wu, Q. (2025). Which agent causes task failures and when? On automated failure attribution of LLM multi-agent systems. In \emph{Proceedings of the 42nd International Conference on Machine Learning}, volume 267 of \emph{Proceedings of Machine Learning Research}, pp. 76583--76599. PMLR.

\end{document}